%% file: main.tex
\documentclass{article}

\PassOptionsToPackage{numbers, compress}{natbib}

\usepackage[preprint]{neurips_2026}

\usepackage[utf8]{inputenc} 
\usepackage[T1]{fontenc}    
\usepackage{hyperref}       
\usepackage{url}            
\usepackage{booktabs}       
\usepackage{amsfonts}       
\usepackage{nicefrac}       
\usepackage{microtype}      
\usepackage{xcolor}         
\input{settings}

\title{Perception Without Engagement: \\Dissecting the Causal Discovery Deficit in LMMs}

\newcommand*\samethanks[1][\value{footnote}]{\footnotemark[#1]}
\author{%
  Jiafeng Liang\textsuperscript{1}\thanks{\ \ Equal contribution}\, , 
  Zhihao Zhu\textsuperscript{1}\samethanks\, , 
  Zihan Zhang\textsuperscript{5}\samethanks\, , 
  Baoqi Ren\textsuperscript{1}, 
  Shixin Jiang\textsuperscript{1},
  Runxuan Liu\textsuperscript{1},\\
  \textbf{Tao Ren\textsuperscript{4}}, 
  \textbf{Ming Liu\textsuperscript{1,2}}\thanks{\ \ Corresponding author}\, , 
  \textbf{See-Kiong Ng\textsuperscript{3}}, 
  \textbf{Bing Qin\textsuperscript{1,2}}\\
  \textsuperscript{1}Harbin Institute of Technology,
  \textsuperscript{2}Pengcheng Laboratory, 
  \textsuperscript{3}National University of Singapore,\\
  \textsuperscript{4}Peking University 
  \textsuperscript{5}Harvard University \\
  \texttt{\{jfliang, zhzhu, mliu\}@ir.hit.edu.cn} \\
  \vspace{-30pt}
}

\begin{document}

\maketitle
\input{sec/0_abstract}
\input{sec/1_introduction}
\input{sec/6_related_work}
\input{sec/2_procaueval}
\input{sec/3_evaluation}
\input{sec/4_adpo}
\input{sec/5_experiment}
\input{sec/7_conclusion}


\bibliography{natbib}
\bibliographystyle{unsrtnat}

\input{sec/8_appendix}


\end{document}

%% file: settings.tex
\usepackage{graphicx} 
\usepackage{algpseudocode}
\usepackage{textcomp}
\usepackage{amssymb}
\title{pseudocode}
\usepackage{CJKutf8}
\usepackage{bbm} 

\usepackage{times}
\usepackage{latexsym}
\usepackage{graphicx}
\usepackage{makecell}
\usepackage{colortbl}
\usepackage{pifont}

\usepackage[T1]{fontenc}
\usepackage{xcolor}
\usepackage[utf8]{inputenc}

\usepackage{microtype}

\usepackage{pifont}
\usepackage{tabularx}
\usepackage{adjustbox}
\usepackage{multirow}
\usepackage{enumitem}
\usepackage{xspace}
\usepackage{tcolorbox}
\usepackage{xparse}
\usepackage{soul}
\usepackage{booktabs,dcolumn}
\usepackage{url}
\usepackage{amsthm,amssymb,bm,stmaryrd}
\usepackage{breqn}
\usepackage{wrapfig}
\usepackage{arydshln}

\usepackage{arydshln} 
\usepackage{makecell}
\usepackage{booktabs}
\usepackage{amsmath}
\usepackage{color}
\usepackage[table]{xcolor}
\definecolor{yellow_l}{RGB}{191,144,0}
\definecolor{blue_l}{RGB}{0,176,240}
\definecolor{blue_case}{RGB}{16,111,183}
\definecolor{yellow_case}{RGB}{255,230,153}
\definecolor{colored_case1}{RGB}{0,86,255}
\definecolor{colored_case2}{RGB}{255,0,0}
\definecolor{colored_case3}{RGB}{154,243,0}
\definecolor{colored_case4}{RGB}{248,115,23}
\definecolor{colored_case5}{RGB}{178,74,255}
\definecolor{colored_case6}{RGB}{255,154,212}
\definecolor{colored_case7}{RGB}{127,96,0}
\usepackage{multirow}
\usepackage{enumitem}
\usepackage{bbding}
\usepackage{balance}
\usepackage{flushend}

\definecolor{a}{RGB}{251, 229, 214}
\definecolor{b}{RGB}{222, 235, 247}
\definecolor{c}{RGB}{226, 240, 217}

\definecolor{hui}{RGB}{242, 242, 242}
\definecolor{lv}{RGB}{210, 255, 218}

\definecolor{lan}{RGB}{30, 119, 179}
\definecolor{zi}{RGB}{112, 48, 159}
\definecolor{cheng}{RGB}{255, 127, 15}

\usepackage{tcolorbox}
\usepackage{xcolor}
\tcbuselibrary{breakable, skins}

\tcbset{
  promptbox/.style={
    breakable,
    enhanced,
    colback=gray!5!white,
    colframe=gray!40!black,
    colbacktitle=gray!20!white,
    coltitle=black,
    fonttitle=\small\bfseries,
    title={#1},
    boxrule=0.5pt,
    arc=4pt,
    left=6pt, right=6pt, top=4pt, bottom=4pt,
    attach boxed title to top left={yshift=-2mm, xshift=6pt},
    boxed title style={
      colback=gray!20!white,
      colframe=gray!40!black,
      boxrule=0.5pt,
      arc=3pt,
    },
  }
}

%% file: sec/0_abstract.tex
\begin{abstract}
Although Large Multimodal Models (LMMs) have achieved strong performance on general video understanding, their susceptibility to textual prior shortcuts during causal discovery has been recognized as a critical deficit.
The underlying mechanisms of this phenomenon remain incompletely understood, as existing benchmarks only measure response accuracy without revealing the sources and extent of the deficit.
We introduce \textbf{\textsc{ProCauEval}}, a perturbation-based evaluation protocol that shifts from outcome assessment to mechanism diagnosis, probing causal discovery through five controlled configurations that systematically manipulate visual and textual modalities to decompose their respective contributions to model behavior and dissect the failure modes.
Evaluating 17 mainstream LMMs, we find that models faithfully perceive video content yet systematically underexploit it during causal reasoning.
We further observe that stronger post-training amplifies rather than mitigates textual prior reliance, and that higher performance correlates with greater fragility under perturbation.
To address these, we propose \textbf{A}nti-\textbf{D}istillation \textbf{P}olicy \textbf{O}ptimization (\textbf{ADPO}), a reinforcement learning framework built on negative teacher alignment, which augments GRPO by explicitly pushing the policy away from a prior-only counterfactual teacher induced by visual corruption. 
Specifically, ADPO maximizes the divergence between the policy distributions conditioned on the original and visually corrupted inputs, thereby forcing the model to ground its reasoning in visual evidence rather than textual shortcuts.
Extensive experiments show that ADPO improves visual engagement without sacrificing fundamental comprehension, thus offering a preliminary step toward reliable causal discovery.
\end{abstract}

%% file: sec/1_introduction.tex
\section{Introduction}

Large Multimodal Models (LMMs)~\citep{llavaov, qwen3vl, internvl35} have achieved remarkable progress across a wide range of video understanding tasks such as action recognition~\citep{kinetics}, visual question answering~\citep{nextqa}, and scene description~\citep{msrvtt}. 
Yet in real-world scenarios, video understanding demands more than perceiving what is visible, requiring instead the ability to reason about why observed events occur, such as inferring why a vehicle brakes suddenly in autonomous driving or what triggers an abnormal tissue response in medical video analysis.
These settings share a common demand for causal discovery~\citep{mecd+}, namely identifying directed cause-effect relationships between events, a qualitatively harder challenge than the perceptual recognition tasks dominating current benchmarks.
Given this gap, whether LMMs genuinely possess robust causal discovery capability remains an open and pressing question.

Recently, causal video benchmarks~\citep{causalstep, mecd+} have made notable progress, most evaluate causal discovery exclusively through outcome accuracy, measuring \emph{how well} a model performs without investigating \emph{why} it fails or \emph{to what extent} each failure source contributes to the overall deficit.
Some benchmarks~\citep{causalvqa, vcrbench} have recognized the risk of statistical shortcut leveraging, where models arrive at correct answers by relying on superficial cues such as linguistic patterns or knowledge biases rather than genuinely reasoning over visual content. 
To mitigate this, they attempt to remove exploitable cues from the evaluation data, whether by filtering text-solvable questions or avoiding predictable answer formats.
However, such a defensive strategy faces an inevitable arms race, as increasingly capable models inevitably discover subtler statistical regularities that benchmark designers did not anticipate.

To address this gap, we introduce \textbf{\textsc{ProCauEval}}, a perturbation-based evaluation protocol that shifts the paradigm from outcome evaluation to mechanism dissection.
Rather than eliminating shortcuts, \textsc{ProCauEval} deliberately injects them through controlled perturbations and examines whether models still ground their reasoning in visual evidence.
Specifically, \textsc{ProCauEval} defines five controlled configurations, namely Veridicality, Plausibility, Incongruence, Deprivation, and Diagnosis, which jointly perturb both visual and textual modalities, enabling a causal decomposition of each modality's contribution to model behavior.
Complemented by five evaluation metrics that capture flip, non-flip, and perceptual fidelity beyond answer accuracy, the protocol provides a fine-grained characterization of the extent and mechanisms of failures in model causal discovery.

Using \textsc{ProCauEval}, we conduct a systematic evaluation of 17 mainstream LMMs spanning instruction-tuned, reasoning-oriented, and large-parameter variants, and arrive at a set of findings. 
First, beyond the known risk of textual shortcut exploitation, our controlled decomposition pinpoints the failure occurring at the reasoning stage rather than perception, as models faithfully comprehend video content yet systematically discard it when inferring causal relations. 
Second, reasoning-oriented post-training~\citep{grpo} amplifies rather than mitigates this dependence on textual priors and may even render visual engagement counterproductive, as models perceive video content yet are misled rather than informed by it during causal reasoning. 
Third, higher baseline accuracy correlates strongly with greater fragility under perturbation, revealing that top-performing models are precisely those whose apparent causal discovery is most unsubstantiated.
These findings collectively locate the core deficit not in visual perception but in cross-modal integration and reasoning, where visual signals are systematically suppressed by textual priors within the thinking process.

\begin{wrapfigure}{r}{0.31\linewidth}
    \vspace{-0.45cm}
    \centering
    \includegraphics[width=\linewidth]{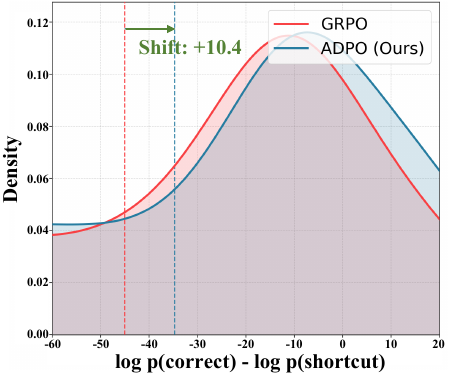}
    \caption{Correct and shortcut output likelihood distribution gap between GRPO and our proposed ADPO.}
    \label{fig:distribution}
    \vspace{-0.4cm}
\end{wrapfigure}

Building on this dissection, we further propose \textbf{A}nti-\textbf{D}istillation \textbf{P}olicy \textbf{O}ptimization (\textbf{ADPO}), a reinforcement learning framework that directly targets the prior-dependent failure mode exposed by \textsc{ProCauEval}. 
In contrast to conventional distillation~\citep{distill}, which pulls a student toward a teacher, ADPO pushes the policy away from a degenerate prior-only teacher that has been deprived of visual input. 
Specifically, built upon Group Relative Policy Optimization (GRPO), ADPO introduces a perception-sensitivity regularizer that maximizes the divergence between the policy distributions conditioned on the original video and on a visually corrupted counterpart, thereby forcing the model to ground its reasoning in visual evidence. 
As shown in Figure~\ref{fig:distribution}, ADPO significantly increases the model's likelihood preference for correct answers over prior shortcut, steering models toward robust causal discovery more firmly grounded in visual evidence.
Our main contributions are as follows:
\begin{itemize}[leftmargin=*]
\item We design \textsc{ProCauEval}, a perturbation-based evaluation protocol with five controlled configurations, shifting causal evaluation from outcome accuracy to mechanism diagnosis.
\item We evaluate 17 mainstream LMMs and reveal that their causal discovery is largely spurious, with the failure located in cross-modal integration and reasoning rather than perception.
\item We propose Anti-Distillation Policy Optimization (ADPO), a reinforcement learning framework that augments GRPO with a perception-sensitivity regularizer to improve model visual engagement.
\end{itemize}
%

%% file: sec/6_related_work.tex
\section{Related Work}

\textbf{Causal Discovery Benchmarks.}
Evaluating causal discovery in video models has attracted growing attention~\citep{nextqa,casualvidqa,nextgqa}. 
MECD~\citep{mecd+} formulates event-level causal graph discovery over videos, and CausalStep~\citep{causalstep} extends this to stepwise causal chain reasoning.
CausalVQA~\citep{causalvqa} and VCRBench~\citep{vcrbench} further probe causal capabilities through physically grounded and long-form video settings, respectively. 
A recurring concern in these benchmarks is shortcut exploitation, where models rely on statistical cues rather than visual evidence.
Existing fixes such as filtering text-solvable questions, designing harder distractors, or balancing answers are defensive patches that struggle to prevent stronger models from capturing subtle regularities.
Our \textsc{ProCauEval} instead injects controlled perturbations to diagnose failure mechanisms, shifting from outcome evaluation to failure dissection.

\textbf{Group Relative Policy Optimization for Video Reasoning.}
Group Relative Policy Optimization (GRPO)~\citep{grpo} has become the dominant reinforcement learning framework for post-training LMMs~\citep{visionr1,r1vl,r1ov,vlmr1,r1reward}. 
For video reasoning, most existing methods simply extend GRPO with visual inputs and directly transfer the algorithm to video tasks~\citep{longvilar1,glm41v,onethinker,keye1.5} or optionally augmenting it with auxiliary task-specific scalar rewards~\citep{videor1,videochatr15,mimovl}.
Although these methods improve chain-of-thought quality, our analysis indicates that they paradoxically amplify reliance on textual priors, as the verifier rewards answer correctness without distinguishing whether it arises from visual grounding or textual co-occurrence. 
Departing from conventional distillation that pulls a student toward a positive teacher~\citep{distill}, our ADPO inverts this paradigm by treating a visually-deprived policy as a negative teacher and repelling the grounded distribution away from it via asymmetric stop-gradient KL.

%% file: sec/2_procaueval.tex
\section{Causal Evaluation Protocol (\textsc{ProCauEval})}
We propose a causal evaluation protocol (\textsc{ProCauEval}), covering framework, data, and metrics, shifting the evaluation paradigm from outcome accuracy to mechanism dissection. 
The core idea is to actively inject controlled perturbations and examine whether model judgments shift in response, thereby quantitatively characterizing the sources and extent of failures in causal discovery.

\subsection{Causal Discovery in Video Comprehension}
\label{Causal Discovery}
Causal discovery~\citep{mecd+} refers to the task of identifying causal relationships among chronologically ordered events within a video, with the goal of constructing a structured, event-level causal diagram that explains why the final event occurred. 
Formally, given a video $V$ accompanied by an event set $E=\left \{ e_{1}^{ca}, e_{2}^{ca},... ,e_{n-1}^{ca}, e_{n}^{ef} \right \}$ where all events correspond to occurrences within the video. 
The final event $E_{e} =\left \{ e_{n}^{ef}\right \}$ is designated as the effect event, while the remaining candidate events are partitioned into two subsets based on a binary label indicating whether an event constitutes a cause.
Specifically, the cause event set $E_{c}=\left \{e_{j}^{ca} \mid \text{label}(e_j^{ca}) = 1\right \}$ contains events labeled as causes, whereas the non-cause event set $E_{o}=\left \{e_{k}^{ca} \mid \text{label}(e_k^{ca}) = 0\right \}$ contains events negatively labeled.

\subsection{Evaluation Framework}
\label{Evaluation Setting}
To conduct a comprehensive ablation and comparative analysis, we systematically design five distinct data pairing configurations. 
Specifically, by selectively modifying the input content of a given modality, we are able to progressively disentangle the causal influence of each modality on the LMMs, and identify the underlying reasons for the failure. See Appendix~\ref{app:evaluation settings} for more details.

\subsubsection{Data Elements}
\begin{figure*}[t]
    \centering
    \includegraphics[width= \linewidth]{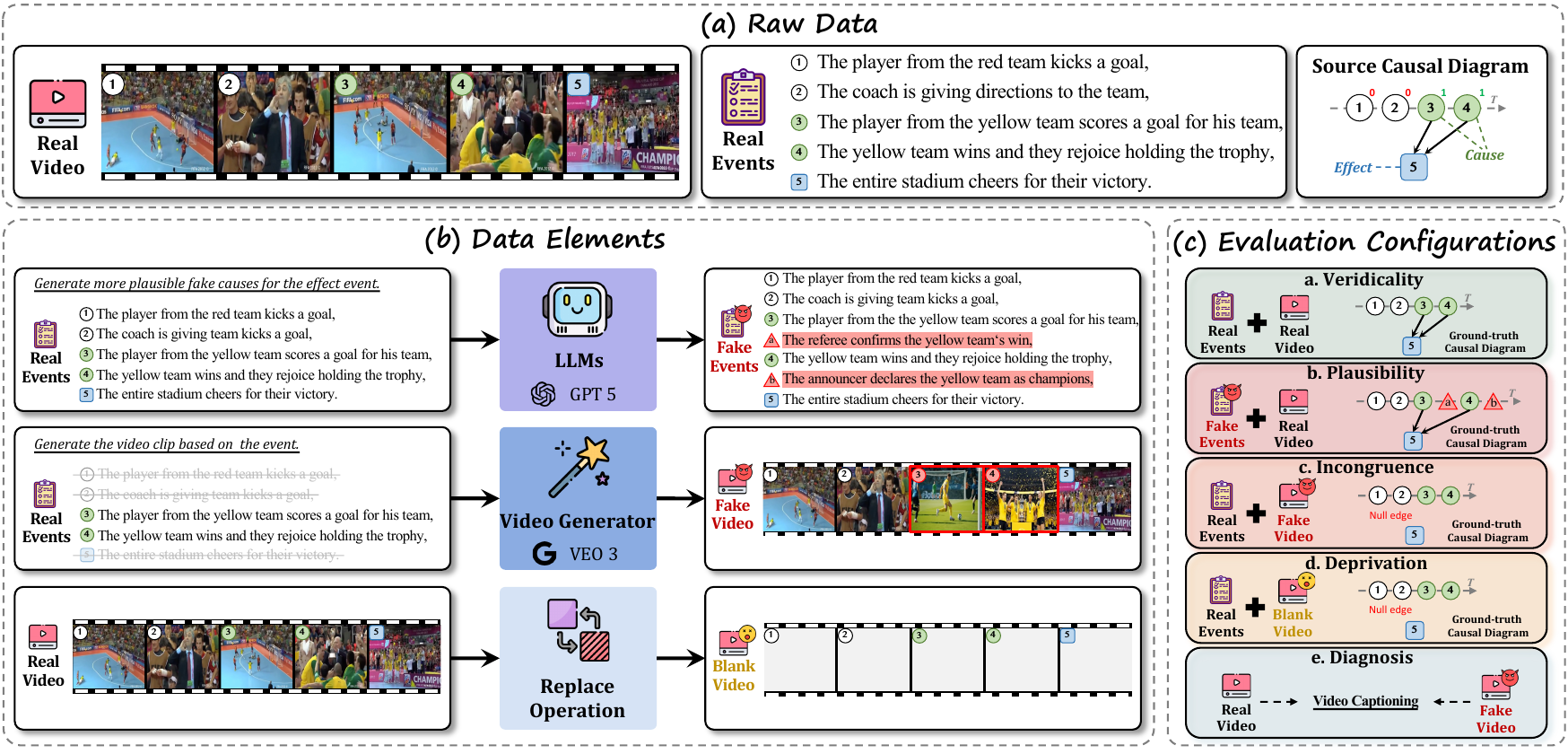}
    \caption{Overview of \textsc{ProCauEval} framework covering data elements and configurations.}
    \label{fig:evaluation}
\end{figure*}

We first construct the data elements (shown in Figure~\ref{fig:evaluation} (b)), which serve as the fundamental building blocks for the evaluation configurations described below. 
The data elements are defined as:

\begin{itemize}[leftmargin=*]
	\item \textbf{\textit{a. Real Events}:} Events that actually occurred in the video and are described in textual form.
    \item \textbf{\textit{b. Real Video}:} Video that authentically contain multiple causal events (i.e., video clip form). 
    \item \textbf{\textit{c. \textcolor{red}{Fake Events}}:}
    The textual event set is constructed by inserting fabricated cause events $E_{f}=\left \{e_{l}^{ca}\right \}$ into the original event sequence. We use GPT-5 \citep{gpt5} to generate candidate fabricated causes and retain those with higher language-only log-probability than the ground-truth causes, making them statistically plausible distractors rather than model-specific adversarial examples.
    \item \textbf{\textit{d. \textcolor{red}{Fake Video}}:}
    Constructed by replacing cause event clips in real videos with fabricated ones generated by Google VEO-3~\citep{veo3}, conditioned on the corresponding cause event text descriptions. 
    By design, generation is conditioned solely on event semantics, yielding clips that are semantically faithful yet scene-inconsistent with surrounding context, a deliberate property probing cross-modal integration. 
    Low-quality clips are filtered via CLIP similarity~\citep{clip} thresholding.
    \item \textbf{\textit{e. \textcolor{yellow_l}{Blank Video}}:}
    All event clips are replaced with blank segments of white frames.  
\end{itemize}

\subsubsection{Evaluation Configurations}
To comprehensively assess the model's causal discovery capability, underlying mechanisms and potential failure modes, we establish five evaluation configurations (shown in Figure~\ref{fig:evaluation} (c)). 
\begin{itemize}[leftmargin=*]
	\item \textbf{\textit{a. \texttt{Veridicality} (Real Events + Real Video)}:}
    Primarily serves to verify whether the model possesses causal capability. 
    Rather than examining the source or reliability of the model's causal discovery, it focuses solely on assessing whether correct judgments can be made.
    \item \textbf{\textit{b. \texttt{\textbf{Plausibility}} (\textcolor{red}{Fake Events} + Real Video)}:}
    This configuration investigates whether the model relies on statistical correlations derived from textual priors alone (i.e., pre-trained knowledge rather than genuinely reasoning over the provided content) when performing causal discovery.
    \item \textbf{\textit{c. \texttt{Incongruence} (Real Events + \textcolor{red}{Fake Video})}: }
    To examine whether the model performs causal discovery by jointly reasoning over the full visual context and aligning it with the textual events.
    \item \textbf{\textit{d. \texttt{Deprivation} (Real Events + \textcolor{yellow_l}{Blank Video})}:}
    This configuration introduces a single-variable difference in the video modality relative to \texttt{Veridicality}, enabling controlled comparison. 
    \item \textbf{\textit{e. \texttt{Diagnosis} (Real and Fake Video Captioning)}:}
    To assess general video understanding independently of causal discovery, captions are generated from both original and fabricated cause videos clips. 
    This rules out the possibility that observed failures in causal discovery reflect an inability to comprehend video content rather than genuine deficiencies in causal reasoning.
    
\end{itemize}

\input{tab/statics}

\subsection{Data Statistics}
\label{Benchmark Statistics}
Our evaluation data is built upon the causal discovery benchmark MECD~\citep{mecd+}. 
We collect 500 videos and automatically construct 2,000 QA pairs based on the original annotations. 
Each video is associated with four distinct evaluation settings based on inconsistency perturbations. 
Furthermore, to ensure the comprehensiveness of the evaluation, we deliberately select videos of varying durations. 
Details are presented in Table~\ref{tab:data_statistics}.

\subsection{Evaluation Metrics}
\label{Evaluation Metrics}
In this section, we introduce the five proposed metrics used in our evaluation. 
We begin with the general metrics that are applicable to most evaluation configurations:

\textbf{Average Score (R-Avg):}
R-Avg measures the average alignment between predicted cause events and cause events in the source causal diagram.
For sample $i$, given the set of cause events $E_{c}$ from the source causal diagram and the set of predicted cause events $\hat{E}_{c}$, R-Avg is defined as:
\begin{equation}
    \mathrm{R\text{-}Avg}=\frac{1}{D} \sum_{i=1}^{D}\frac{\left | E_{c}\cap \hat{E} _{c} \right | }{\left | E_{c} \right | } ,
\end{equation}
where $D$ denotes the total number of test samples, $|\cdot|$ denotes the cardinality of a set.
Furthermore, we introduce four additional proposed metrics, each tailored to specific evaluation configurations:

\textbf{Accuracy (Acc):}
Acc is specifically designed for \texttt{Incongruence} and \texttt{Deprivation}, in which the ground-truth causal diagram contains no causal edges. 
To accommodate this, we introduce an additional candidate option $E_{n} = \{\text{``}\mathrm{Based \\\ on\\\ this\\\ video, none\\\ of\\\ them} \text{''}\}$, where selecting this option is the only correct answer, formulated as:
\vspace{-4pt}
\begin{equation}
    \mathrm{Acc} = \frac{1}{D}\sum_{i=1}^{D} \mathbf{1}\left[\hat{E}_{c}=E_{n}\right].
\end{equation}

\textbf{Flip Rate (FR):}
FR is proposed for configuration \texttt{Plausibility} to assess how susceptible the model is to statistically-biased textual perturbations. 
Specifically, FR measures the proportion of responses that were correct under configuration \texttt{Veridicality} but were flipped to match the fabricated events introduced in configuration \texttt{Plausibility}, formulated as:
\begin{equation}
    \mathrm{FR} = \frac{\sum_{i=1}^{D} \mathbf{1_{_V}}\left[\frac{\left | E_{c}\cap \hat{E}_{c} \right | }{\left | E_{c} \right | }=1\right] \cdot  \mathbf{1_{_P}}\left[\frac{\left | E_{f} \cap \hat{E}_{c} \right | }{\left | E_{c} \right | }=1\right]}{\sum_{i=1}^{D} \mathbf{1_{_V}}\left[\frac{\left | E_{c}\cap \hat{E}_{c} \right | }{\left | E_{c} \right | }=1\right]},
\end{equation}
where $\mathbf{1_{_V}}[\cdot]$ and $\mathbf{1_{_P}}[\cdot]$ are indicator functions under configurations \texttt{Veridicality} and \texttt{Plausibility}, respectively.
A higher FR indicates that the model is more easily misled into selecting fabricated events over the original correct causes.

\textbf{Non-Flip Rate (NFR):}
When video content is replaced, an ideal model should update its causal judgments accordingly. 
Specifically, NFR measures the proportion of responses that were correct under \texttt{Veridicality} and remained unchanged after the video was manipulated in configurations \texttt{Incongruence} or \texttt{Deprivation}.
\begin{equation}
    \mathrm{NFR} = \frac{\sum_{i=1}^{D} \mathbf{1_{_V}}\left[\frac{\left | E_{c}\cap \hat{E}_{c} \right | }{\left | E_{c} \right | }=1\right] \cdot  \mathbf{1_{I,D}}\left[\frac{\left | E_{c} \cap \hat{E}_{c} \right | }{\left | E_{c} \right | }=1\right]}{\sum_{i=1}^{D} \mathbf{1_{_V}}\left[\frac{\left | E_{c}\cap \hat{E}_{c} \right | }{\left | E_{c} \right | }=1\right]},
\end{equation}
where $\mathbf{1_{I,D}}[\cdot]$ means indicator functions under configurations \texttt{Incongruence} or \texttt{Deprivation}.
A higher NFR exposes the model's failure to leverage visual information for causal reasoning.

\textbf{GPT Score:}
GPT Score is introduced for configuration \texttt{Diagnosis} to evaluate the fine-grained quality of video comprehension. 
We design a detailed prompt that assesses each caption from multiple dimensions, assigning score on scale of 1 to 10. 
See Appendix~\ref{app:evaluation settings} for evaluation prompt.

%% file: tab/statics.tex
\begin{wraptable}{r}{0.3\linewidth}
\centering
\small
\setlength{\tabcolsep}{6pt}
\renewcommand\arraystretch{1.10}
\vspace{-1.5cm}
\caption{Data statistics}
\vspace{4pt}
\label{tab:data_statistics}
\begin{tabular}{@{}lr@{}}
    \toprule
    \textbf{Category} & \textbf{Size} \\
    \midrule
    Video Sources & 500 \\
    ~- Avg events per Video & 5.1 \\
    ~- Max Duration & 536.9s \\
    ~- Min Duration & 8.9s \\
    ~- Avg Duration & 132.8s \\
    \midrule
    Questions per Video & 4 \\
    Total Samples & 2000 \\
    \bottomrule
\end{tabular}
\vspace{-1.0cm}
\end{wraptable}

%% file: sec/3_evaluation.tex
\section{Evaluations}

\subsection{Models}
\label{Models}
We conduct a comprehensive evaluation of most existing 17 mainstream LMMs, encompassing both standard instruction-tuned models and reasoning-oriented thinking models. 
Additionally, we extend our evaluation to include models of larger parameter scales. See Appendix~\ref{app:evaluation settings} for more details.

\subsection{Results and Analysis}
\label{Results and Analysis}
\input{tab/evaluation}
\textbf{\begin{figure*}[t]
    \centering
    \includegraphics[width= \linewidth]{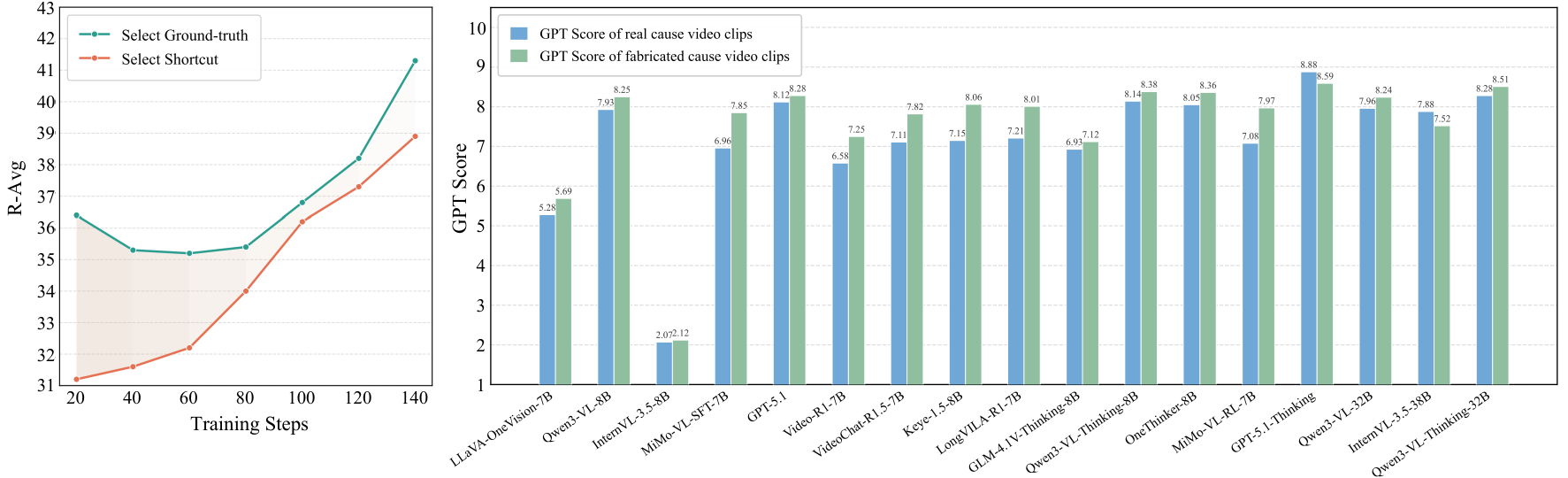}
    \caption{(a) Selection trends of the model toward correct and shortcut answers during GRPO training. (b) GPT Scores of LMMs' perception capability on real and fabricated cause video clips.}
    \vspace{-14pt}
    \label{fig:posttrain_gptscore}
\end{figure*}}
\textbf{\textit{Videos are Substantially Underused in Causal Discovery.}}
A central finding from Table~\ref{tab:evaluation} is that video information is largely bypassed during causal discovery.
The most direct evidence comes from the \texttt{Plausibility}, in which models that genuinely ground their judgments in visual content would be expected to maintain a stable R-Avg. 
However, the mean R-Avg across all 17 LMMs collapses from $33.9\%$ under \texttt{Veridicality} to merely $2.4\%$, corresponding to an average relative drop of $88.7\%$. 
This conclusion is reinforced from the opposite direction by the \texttt{Incongruence} and \texttt{Deprivation}, where replacing real videos with fabricated content yields a mean R-Avg difference of only $3.7\%$, and 15 out of 17 models attain an even higher R-Avg when videos are substituted with blank frames. 
Moreover, this deficit is not reliably alleviated by scaling up parameters, as the three scaling pairs in the table exhibit divergent and even inverse trends.
Taken together, these findings suggest that videos do not serve as an effective source of evidence for causal discovery, and models tend to fall back on textual priors, producing judgments that are largely ungrounded.

\textbf{\textit{Fundamental Comprehension and Causal Discovery are Decoupled.}}
A natural concern is whether these failures simply reflect an inability to comprehend video content, a hypothesis directly refuted by the \texttt{Diagnosis}. 
As shown in Figure~\ref{fig:posttrain_gptscore} (b), most LMMs produce captions that faithfully and comprehensively describe both real and fabricated videos, achieving average GPT Scores of $7.3$, indicating reliable perception at the visual level.
The contrast with their behavior under \texttt{Incongruence} is striking, as exemplified by Qwen3-VL-8B-Thinking~\cite{qwen3vl}, which attains a GPT Score of $8.14$ on real videos yet only $8.0\%$ Acc under and a $79.3\%$ NFR under \texttt{Incongruence}. 
This decoupling between perception and evidence utilization supports our central claim that the causal discovery deficit in LMMs is not a perceptual failure but one of cross-modal integration, in which visual signals are systematically suppressed by textual priors.

\textbf{\textit{Post-training Induces an Inverse Scaling between Reasoning Capacity and Visual Engagement.}}
Reasoning-oriented RL with explicit chain-of-thought (CoT), although the dominant post-training paradigm for reasoning, fails to mitigate and even exacerbates the reliance on textual priors. 
As shown in Table~\ref{tab:evaluation}, thinking variants attain marginally higher mean R-Avg under \texttt{Veridicality} than standard variants, yet exhibit substantially higher NFR and lower Acc. 
A controlled evaluation on Qwen3-VL-8B with GRPO further confirms this trend (shown in Figure~\ref{fig:posttrain_gptscore} (a)), as the R-Avg gap between \texttt{Veridicality} and \texttt{Incongruence} narrows progressively across training steps. 
Moreover, several thinking models achieve higher R-Avg under \texttt{Incongruence} than under \texttt{Veridicality}, a pattern largely absent in standard models. 
One plausible explanation is that the extended reasoning chains of thinking models perceive and incorporate rich visual details from real videos, yet lack the capacity to correctly leverage them for causal judgment.
When replaced with scene-inconsistent fabricated content, the visual stream signals a cross-modal conflict, prompting models to fall back to textual priors, paradoxically achieving higher accuracy.
This improvement is essentially ungrounded, as the model is not reasoning better but has disengaged from the video.

\input{tab/pearson}
\textbf{\textit{Higher Accuracy Comes with Greater Fragility.}}
One would intuitively expect that LMMs with higher causal discovery accuracy possess stronger causal capabilities and thus exhibit greater robustness, yet our results reveal precisely the opposite pattern. 
As shown in Table~\ref{tab:pearson}, the Pearson correlation between R-Avg under \texttt{Veridicality} and FR under \texttt{Plausibility} reaches $r = 0.832$ across all 17 LMMs, and similarly strong correlations are observed for NFR under \texttt{Incongruence} and \texttt{Deprivation}, indicating that higher baseline accuracy is consistently associated with greater susceptibility to textual perturbation. 
This paradox is most clearly illustrated by the top-performing LMMs under \texttt{Veridicality}, whose correct predictions are largely flipped to fabricated events under textual perturbation and remain nearly unchanged even when the video is entirely replaced. 
Their high accuracy is therefore not a product of grounded multimodal causal discovery, but rather reflects well-calibrated textual priors that happen to align with the pre-training pattern.

%% file: tab/evaluation.tex
\begin{table*}[t]
	\centering
	\caption{Evaluation results of three types of LMMs under four  configurations.}
	\begin{adjustbox}{width=\textwidth}
	\begin{tabular}{lcccccccccccc}
		\toprule
		\multirow{2.5}{*}{\textbf{Model}} & \multirow{2.5}{*}{\textbf{Size}}  & \multicolumn{1}{c}{\texttt{\textbf{Veridicality}}} & \multicolumn{2}{c}{\texttt{\textbf{Plausibility}}} & \multicolumn{3}{c}{\texttt{\textbf{Incongruence}}} & \multicolumn{3}{c}{\texttt{\textbf{Deprivation}}}
        \\
		\cmidrule(r){3-3}
        \cmidrule(r){4-5}
        \cmidrule(r){6-8}
        \cmidrule(r){9-11}
		& & \textbf{R-Avg} $\uparrow$ & \textbf{R-Avg} $\uparrow$ & \textbf{FR} $\downarrow$ & \textbf{R-Avg} $\downarrow$ & \textbf{Acc} $\uparrow$ & \textbf{NFR} $\downarrow$ & \textbf{R-Avg} $\downarrow$ & \textbf{Acc} $\uparrow$ & \textbf{NFR} $\downarrow$ \\
		\midrule
        \scriptsize \textit{Standard Models} \\
		LLaVA-OneVision~\cite{llavaov} & 7B & 20.4 & 2.6 & 24.7 &20.6 &28.2 &53.3 &23.8 &34.4 &33.8
        \\
		Qwen3-VL~\cite{qwen3vl} & 8B &36.4 &3.9 &51.7 &31.2 &13.2 &54.1 &36.7 &24.6 &52.5
        \\
		InternVL-3.5~\cite{internvl35} & 8B &34.1 &2.7 &30.9 &32.6 &17.0 & 43.1 &50.3 &28.8 &51.2
		\\
        MiMo-VL-SFT~\cite{mimovl} & 7B &34.5 &7.9 &57.9 &33.2 &6.8 &63.2 &35.8 &17.6 &47.4
		\\
        GPT‑5.1~\cite{gpt5} & - &25.8 &1.0 &20.8 &20.5 &28.6 &40.6 &29.3 &23.4 &43.4 
		\\
		\midrule
        \scriptsize \textit{Thinking Models} \\
		Video-R1~\cite{videor1} & 7B &27.9 &4.9 &22.3 &31.6 &16.4 &52.4 &32.8 &26.2 &45.6
		\\
		VideoChat-R1.5~\cite{videochatr15} & 7B &37.1 &11.9 &29.1 &36.6 &17.6 &60.9 &30.3 &19.2 &50.4
        \\
        Keye-1.5~\cite{keye1.5} & 8B &47.8 &3.6 &70.0 &51.2 &14.0 & 78.0 &49.2 &12.0 &73.3
        \\
        LongVILA-R1~\cite{longvilar1} & 7B &20.6 &5.2 &23.0 &18.9 &7.4 &33.8 &25.3 &11.8 &46.0
        \\
        GLM-4.1V-Thinking~\cite{glm41vthinking} & 8B &37.7 &1.1 &36.7 &37.3 &9.8 & 65.5 &43.9 &18.2 &69.1
        \\
        Qwen3-VL-Thinking~\cite{qwen3vl} & 8B &47.6 &3.8 &72.9 &50.1 &8.0 &79.3 &50.2 &8.8 &79.3
        \\
        OneThinker~\cite{onethinker} & 8B &31.2 &2.5 &36.8 &33.1 &4.2 &50.9 &28.3 &20.4 &46.9
        \\
        MiMo-VL-RL~\cite{mimovl} & 7B &30.0 &6.2 &37.8 &34.1 &10.2 &52.4 &38.3 &17.0 &56.1
        \\
        GPT‑5.1-Thinking~\cite{gpt5} & - &33.9 &0.4 &31.9 &38.3 &11.0 &70.2 &39.9 &11.4 &75.2 
        \\
        \midrule
        \scriptsize \textit{Larger Models} \\
        Qwen3-VL~\cite{qwen3vl} & 32B &37.4 &2.6 &37.1 &35.0 &13.4 &68.6 &39.3 &17.2 &62.1 
        \\
        InternVL-3.5~\cite{internvl35} & 38B &26.8 &1.6 &30.2 &29.4 &17.4 &58.2 &43.1 &10.4 &59.3
        \\
        Qwen3-VL-Thinking~\cite{qwen3vl} & 32B &48.0 &3.2 &55.8 &42.4 &8.2 &63.8 &49.3 &22.0 &65.0 
        \\
		
		\bottomrule
	\end{tabular}
\end{adjustbox}
\label{tab:evaluation}
\end{table*}

%% file: tab/pearson.tex
\begin{wraptable}{r}{0.5\linewidth}
\centering
\vspace{-0.675cm}
\caption{Cross-metric correlation analysis revealing the accuracy-fragility paradox. \texttt{V}, \texttt{P}, \texttt{I}, and \texttt{D} denote the \texttt{Veridicality}, \texttt{Plausibility}, \texttt{Incongruence}, and \texttt{Deprivation} configurations, respectively.}
\label{tab:pearson}
\renewcommand{\arraystretch}{1.2}
\setlength{\tabcolsep}{4pt}
\small
\begin{tabular}{lcc}
\toprule
\textbf{Metric Pair} & \makecell{\textbf{Pearson $r$}\\\textbf{(7$\sim$8B)}} & \makecell{\textbf{Pearson $r$}\\\textbf{(All 17)}} \\
\midrule
R-Avg (\texttt{V}) vs. FR (\texttt{P})   & 0.836 & 0.832 \\
R-Avg (\texttt{V}) vs. NFR (\texttt{I})  & 0.824 & 0.774 \\
R-Avg (\texttt{V}) vs. NFR (\texttt{D})  & 0.805 & 0.764 \\
FR (\texttt{P}) vs. NFR (\texttt{I})     & 0.748 & 0.715 \\
\bottomrule
\end{tabular}
\vspace{-0.2cm}
\end{wraptable}

%% file: sec/4_adpo.tex
\section{Anti-Distillation Policy Optimization (ADPO)}
\label{sec:adpo}

LMMs faithfully perceive video content but systematically underuse it during causal discovery, defaulting to textual priors. 
Mainstream rule-based RL methods like Group Relative Policy Optimization (GRPO)~\cite{grpo} struggle to address this, as the verifier only checks answers without distinguishing whether correctness arises from visual understanding or textual co-occurrence, while KL constraint anchors the policy to a biased initialization. 
We propose Anti-Distillation Policy Optimization (ADPO), which introduces a negative teacher alignment paradigm by explicitly repelling the policy from a prior-only counterfactual teacher, forcing the reasoning distribution to ground in visual evidence.

\subsection{Preliminaries: Group Relative Policy Optimization (GRPO)}
\label{sec:adpo_pre}

Given a video-question pair $(V, q)$, the policy $\pi_\theta$ samples a group of $G$ rollouts $\{o_i\}_{i=1}^{G}$, each scored by a rule-based verifier to obtain a scalar reward $r_i$. The group-relative advantage is computed as $A_i = (r_i - \mathrm{mean}(\{r\})) / \mathrm{std}(\{r\})$, and the policy maximizes a clipped surrogate objective regularized by KL divergence toward a reference model $\pi_{\text{ref}}$:
\begin{equation}
\mathcal{J}_{\text{GRPO}}(\theta) = \mathbb{E}\!\left[\frac{1}{G}\sum_{i=1}^{G}\frac{1}{|o_i|}\sum_{t=1}^{|o_i|}\Big[\min\!\big(\rho_{i,t} A_i,\;\mathrm{clip}(\rho_{i,t},1\!-\!\epsilon,1\!+\!\epsilon) A_i\big) - \beta\,\mathbb{D}_{\mathrm{KL}}\!\big[\pi_\theta\|\pi_{\text{ref}}\big]\Big]\right],
\label{eq:grpo}
\end{equation}
where $\rho_{i,t}=\pi_\theta(o_{i,t}\mid V,q,o_{i,<t})/\pi_{\theta_{\text{old}}}(o_{i,t}\mid V,q,o_{i,<t})$ is the importance sampling ratio.

\subsection{Framework of Anti-Distillation Policy Optimization}
\label{sec:adpo_method}

ADPO comprises three stages, namely visual perturbation, perception-sensitivity regularization, and joint optimization. 
Specifically, we apply perturbations to the input video to construct counterfactual views, evaluate the policy on both the original and perturbed inputs respectively, and penalize outputs that are unresponsive to visual changes, thereby suppressing prior-driven shortcut behaviors.

\subsubsection{Visual Perturbation}
For every training instance $(V, q)$, we construct a corrupted counterpart $\tilde{V}$ by applying a stochastic perturbation operator $\mathcal{P}(\cdot)$ to the original video. Specifically, $\mathcal{P}$ is selected from a set of three visual perturbations, namely frame shuffling, spatial cropping, and frame replacement (see Appendix~\ref{app:adpo} for implementation details), each disrupting a distinct aspect of the visual evidence:
\begin{equation}
\tilde{V} \;=\; \mathcal{P}(V), \quad \mathcal{P} \in \{\mathcal{P}_{\text{shuffle}},\, \mathcal{P}_{\text{crop}},\, \mathcal{P}_{\text{replace}}\},
\label{eq:mask}
\end{equation}
where $\mathcal{P}_{\text{shuffle}}$ randomly permutes the temporal order of a fixed number of frames, $\mathcal{P}_{\text{crop}}$ randomly masks a fixed-area region within each frame, and $\mathcal{P}_{\text{replace}}$ replaces a fixed number of randomly selected frames with blank frames. 
The choice of $\mathcal{P}$ is fixed throughout a single training run as a configuration hyperparameter, while its internal randomness is freshly resampled at every optimization step. 
The corrupted video retains enough low-level statistics to remain in-distribution for the vision encoder, while partially disrupting the fine-grained visual evidence that reasoning would otherwise rely upon.

\subsubsection{Perception-Sensitivity Regularization}
At each iteration, ADPO performs two forward passes to recompute the distributions over the same group of rollouts $\{o_i\}_{i=1}^{G}$ generated from the original video, one conditioned on the original visual input $V$ and the other on its perturbed counterpart $\tilde{V}$, yielding two policy distributions respectively:
\begin{equation}
\pi_\theta^{(i,t)} = \pi_\theta(\cdot\mid V, q, o_{i,<t}), \quad
\pi_\theta^{\text{pert},(i,t)} = \pi_\theta(\cdot\mid \tilde{V}, q, o_{i,<t}).
\label{eq:pi_both}
\end{equation}
This ensures that any divergence between $\pi_\theta^{(i,t)}$ and $\pi_\theta^{\text{pert},(i,t)}$ is unambiguously attributable to the model's sensitivity to the visual modality, rather than being confounded by differences in the trajectory. 
We then define the perception-sensitivity regularizer as a repulsion objective against the prior-only counterfactual teacher:
\begin{equation}
\mathcal{L}_{\text{prcp}}(\theta) \;=\; \frac{1}{G}\sum_{i=1}^{G}\frac{1}{|o_i|}\sum_{t=1}^{|o_i|}\mathbb{D}_{\mathrm{KL}}\!\left[\,\pi_\theta^{(i,t)}\,\big\|\,\mathrm{sg}\left (\pi_\theta^{\text{pert},(i,t)}\right) \right],
\label{eq:lprcp}
\end{equation}
where $\mathrm{sg}(\cdot)$ denotes the stop-gradient operator, which enforces asymmetric repulsion from the prior-only counterfactual teacher $\pi_\theta^{\text{pert},(i,t)}$, forcing the grounded distribution $\pi_\theta^{(i,t)}$ to move away from the prior-dominated policy rather than allowing both distributions to collapse toward arbitrary midpoint.

\subsubsection{Joint Optimization}
Combining the standard GRPO surrogate with the perception-sensitivity term, we maximize:
\begin{equation}
\mathcal{J}_{\text{ADPO}}(\theta) \;=\; \mathcal{J}_{\text{GRPO}}(\theta) \;+\; \lambda\,\mathcal{L}_{\text{prcp}}(\theta),
\label{eq:adpo}
\end{equation}
where $\lambda$ is a single hyperparameter controlling the strength of the anti-distillation regularizer. 
The detailed hyperparameter settings analysis are provided in Appendix~\ref{app:adpo}.

%% file: sec/5_experiment.tex
\section{Experiments}
\input{tab/main_results}
\subsection{Experimental Setups}
We compare our ADPO against two RLVR baselines, namely GRPO~\cite{grpo} and DAPO~\cite{dapo}. 
Note that GRPO employs a reference KL penalty, while DAPO removes it and adopts dynamic sampling instead. 
We evaluate the effectiveness of the baselines and our ADPO on two LMMs, namely Qwen3-VL-4B and 8B~\cite{qwen3vl}. 
See Appendix~\ref{app:adpo} for more implementation details.

\subsection{Experiment Result}
Table~\ref{tab:adpo_results} reports the performance of ADPO against competing RLVR baselines. 
ADPO not only attains the strongest overall performance but also exhibits a property absent in both GRPO and DAPO, concurrently enhancing causal discovery accuracy and robustness to perturbation.
While GRPO and DAPO improve \texttt{Veridicality} performance, such gains are accompanied by an intensified dependence on textual priors, as evidenced by elevated FR under \texttt{Plausibility} and higher NFR under \texttt{Incongruence} and \texttt{Deprivation}.
In contrast, ADPO reverses this trade-off, markedly suppressing FR while attaining the highest Acc and lowest NFR across perturbed configurations. 
These findings suggest that conventional RLVR methods achieve apparent accuracy gains by amplifying prior reliance, whereas ADPO explicitly targets and mitigates this failure mode, yielding more reliable multimodal causal discovery grounded in visual evidence.

\subsection{Analysis}
\begin{wrapfigure}{r}{0.449\linewidth}
    \vspace{-0.5cm}
    \centering
    \includegraphics[width=\linewidth]{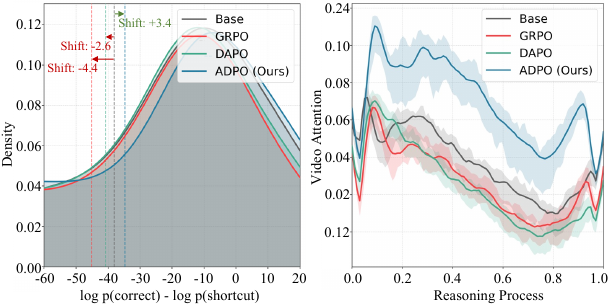}
    \caption{Discriminative capability between correct and shortcut answers measured by log-likelihood gap (left) and attention weights on visual tokens under different methods (right).}
    \label{fig:evidence}
    \vspace{-0.5cm}
\end{wrapfigure}

\textbf{Complementary Evidence of Visual Engagement.}
Figure~\ref{fig:evidence} examines the internal behavior of different strategies from two complementary perspectives. As indirect evidence at the output level, ADPO widens the likelihood gap between correct and shortcut answers, indicating that our method endows the model with a stronger discriminative capability against prior-driven cues. 
As direct evidence at the attention level, ADPO allocates markedly higher weight mass to video tokens during chain-of-thought, demonstrating that it actively restructures the model's internal information flow and channels reasoning more through visual evidence rather than textual priors.

\textbf{Effect of Different Visual Perturbation Strategies.}
As shown in Table~\ref{tab:adpo_results}, all three perturbation strategies consistently outperform the baseline across the four evaluation configurations. 
Among them, frame shuffling yields the most significant improvement, as it disrupts the temporal causal order between events, providing the sharpest contrastive signal for the perception-sensitivity regularizer. 
These results indicate that temporal dynamics constitute a critical dimension of video information, and disrupting the temporal order is the most effective way to corrupt video evidence.

\input{tab/general_video}
\textbf{General Video Understanding Capability.}
A natural concern is whether the perception-sensitivity regularizer, while improving causal discovery, compromises the model's general video understanding capability. 
To address this, we evaluate ADPO on a suite of standard video benchmarks, namely Video-MME~\citep{videomme}, LongVideoBench~\cite{longvideobench}, MMVU~\cite{mmvu}, and MVBench~\cite{mvbench}, covering perception and reasoning over different video durations.
As shown in Table~\ref{tab:general_capability}, ADPO achieves better across all benchmarks, indicating that the anti-distillation objective does not come at the cost of fundamental comprehension. 

%% file: tab/main_results.tex
\begin{table*}[t]
	\centering
	\caption{Experimental results of ADPO compared with baselines across different backbones. We additionally report three variants corresponding to different visual perturbation operators $\mathcal{P}$.}
	\begin{adjustbox}{width=\textwidth}
	\begin{tabular}{llcccccccccc}
		\toprule
		\multirow{2.5}{*}{\textbf{Backbone}} & \multirow{2.5}{*}{\textbf{Method}}  & \multicolumn{1}{c}{\texttt{\textbf{Veridicality}}} & \multicolumn{2}{c}{\texttt{\textbf{Plausibility}}} & \multicolumn{3}{c}{\texttt{\textbf{Incongruence}}} & \multicolumn{3}{c}{\texttt{\textbf{Deprivation}}}
        \\
		\cmidrule(r){3-3}
        \cmidrule(r){4-5}
        \cmidrule(r){6-8}
        \cmidrule(r){9-11}
		& & \textbf{R-Avg} $\uparrow$ & \textbf{R-Avg} $\uparrow$ & \textbf{FR} $\downarrow$ & \textbf{R-Avg} $\downarrow$ & \textbf{Acc} $\uparrow$ & \textbf{NFR} $\downarrow$ & \textbf{R-Avg} $\downarrow$ & \textbf{Acc} $\uparrow$ & \textbf{NFR} $\downarrow$ \\
		\midrule
		\multirow{4}{*}{\textit{Qwen3-VL-4B}}
		& Base~\citep{qwen3vl}    & 30.2 & 1.7 & 33.3 & 29.7 & 11.4 & 50.4 & 26.5 & 35.0 & 36.7 \\
		& GRPO~\citep{grpo}    & 32.8 & 2.9 & 39.5 & 38.8 & 8.6 & 59.7 & 28.7 & 29.6 & 41.0 \\
		& DAPO~\citep{dapo}    & 34.1 & 1.8 & 43.5 & 33.1 & 12.0 & 53.6 & 21.6 & 41.2 & 27.5 \\
		& \textbf{ADPO} (Ours)  & \textbf{36.4} & \textbf{3.4} & \textbf{25.7} & \textbf{28.1} & \textbf{14.0} & \textbf{39.0} & \textbf{17.5} & \textbf{47.4} & \textbf{26.0} \\
		\midrule
		\multirow{6}{*}{\textit{Qwen3-VL-8B}} 
		& Base~\citep{qwen3vl}    & 36.4 & 3.9 & 51.7 & 31.2 & 13.2 & 54.1 & 36.7 & 24.6 & 52.5 \\
		& GRPO~\citep{grpo}    & 41.3 & 4.2 & 52.1 & 38.9 & 10.0 & 60.6 & 38.4 & 20.8 & 55.9 \\
		& DAPO~\citep{dapo}    & 39.2 & 2.6 & 48.9 & 39.0 & 9.4  & 54.3 & 37.9 & 20.8 & 58.3 \\
		& \textbf{ADPO w/ \textit{Shuffle}} (Ours)  & \textbf{43.1} & \textbf{4.2} & \textbf{42.2} & \textbf{30.9} & \textbf{17.2} & \textbf{47.0} & \textbf{33.7} & \textbf{32.0} & \textbf{43.1} \\
		& \textbf{ADPO w/ \textit{Crop}}     & 41.9 & 3.9 & 46.3 & 33.9 & 15.0 & 53.9 & 37.1 & 25.0 & 47.3 \\
		& \textbf{ADPO w/ \textit{Replace}}  & 41.4 & 4.0 & 45.5 & 31.6 & 14.4 & 51.2 & 36.6 & 27.0 & 49.3 \\
		\bottomrule
	\end{tabular}
\end{adjustbox}
\label{tab:adpo_results}
\vspace{-0.4cm}
\end{table*}

%% file: tab/general_video.tex
\begin{wraptable}{r}{0.6\textwidth}
	\centering
	\vspace{-0.65cm}
	\caption{General video understanding performance.}
	\begin{adjustbox}{width=0.6\textwidth}
	\begin{tabular}{lcccc}
		\toprule
		\textbf{Method} & \textbf{Video-MME} & \textbf{LongVideoBench} & \textbf{MMVU} & \textbf{MVBench} \\
		\midrule
		\textit{Qwen3-VL-8B} & 64.7 & 56.3 & 69.7 & 65.3 \\
		w/ GRPO & 65.4 & 60.5 & 70.2 & 66.0 \\
		w/ DAPO & 65.9 & 60.1 & 72.0 & 67.1 \\
		w/ \textbf{ADPO} (Ours) & \textbf{66.7} & \textbf{62.3} & \textbf{73.1} & \textbf{68.7} \\
		\bottomrule
	\end{tabular}
	\end{adjustbox}
	\label{tab:general_capability}
	\vspace{-0.35cm}
\end{wraptable}

%% file: sec/7_conclusion.tex
\section{Conclusion}
In this paper, we shift the evaluation of causal discovery in Large Multimodal Models (LMMs) from outcome accuracy to mechanism diagnosis through \textsc{ProCauEval}, a perturbation-based protocol that decomposes each modality's contribution via five controlled configurations.
Evaluating 17 mainstream LMMs, we find that visual comprehension is preserved while causal reasoning bypasses it for textual priors, and this deficit amplified by reasoning-oriented post-training and most pronounced in top-performing models.
Building on this diagnosis, we propose Anti-Distillation Policy Optimization (ADPO), which introduce the prior-only distribution as a negative teacher augments GRPO with a perception-sensitivity regularizer that maximizes divergence between policy distributions under original and visually corrupted inputs. 
Extensive experiments demonstrate that ADPO grounds reasoning in visual evidence, offering a preliminary step toward reliable multimodal causal discovery.

%% file: sec/8_appendix.tex
\appendix

\section{Further Discussion, Limitations, and Future Work}
\label{app:Limitation}
While our work introduces \textsc{ProCauEval} and ADPO with promising results, we acknowledge several limitations that suggest avenues for future work:

\textbf{Language coverage.} Our evaluation is conducted exclusively in English. 
Although the proposed protocol is language-agnostic in principle, extending \textsc{ProCauEval} to multilingual settings, particularly low-resource languages, remains an interesting direction for future investigation.

\textbf{Video duration distribution.} The videos in our benchmark range from 8.9s to 536.9s, with an average duration of 132.8s. While this range covers a reasonable spectrum of typical video content, ultra-long videos (e.g., feature-length films exceeding one hour) are not represented in our current evaluation set. 
Whether the observed causal discovery deficit persists at substantially longer temporal scales remains to be verified.

\textbf{Frame sampling rate during training.} Due to computational constraints, we uniformly sample 16 frames per video during ADPO training. 
Although we use 128 frames at evaluation time to ensure fair comparison, exploring denser frame sampling during training could potentially yield further improvements.

\textbf{Evaluation cost of GPT Score.} The \texttt{Diagnosis} configuration relies on GPT-based scoring for caption quality assessment. 
While this provides a fine-grained evaluation, it incurs API costs and introduces a dependency on a closed-source model. Future work could explore open-source alternatives for caption quality assessment.

\section{Evaluation Settings}
\label{app:evaluation settings}

\subsection{Fake Events Construction and Potential Leakage Concern}
Fake events are designed as controlled textual perturbations for diagnosing whether the model genuinely relies on visual evidence during causal discovery. 
Rather than serving as ordinary distractors, they are constructed to be plausible shortcut candidates under a prior-only inference setting.
Specifically, we retain fabricated causes that receive higher language-only log-probability than the original causes, ensuring that the model would be likely to select them if visual evidence were not effectively used. 
This allows \texttt{Plausibility} to distinguish reliable visual grounding from prior-driven reasoning: If the shortcut is already attractive under text-only inference, resisting it under multimodal input provides stronger evidence that the model is reasoning over visual content.

\subsection{Fairness of Model-Specific Fake Event Selection}
A potential concern is that selecting fake events based on each model's own log-probabilities could yield different perturbation sets across models, raising questions about cross-model comparability.
To eliminate this concern, we adopt a unified fake event set constructed as the intersection of model-preferred candidates across all evaluated LMMs.
Specifically, for each video, we retain only those fabricated cause events that receive higher language-only log-probability than the ground-truth causes under every model in our evaluation pool, ensuring that all models face an identical set of perturbations.
This intersection-based design guarantees strict cross-model comparability: 
Every model is evaluated against the same pool of fake events, eliminating confound introduced by model-specific selection.
Moreover, an event surviving this intersection is, by construction, a statistically tempting shortcut for all models simultaneously, which makes it a stronger and more universal probe of textual prior reliance than any per-model selection.
Consequently, the FR metric reported in our evaluation reflects a uniform and rigorous measurement of shortcut dependence that is directly comparable across the entire model suite.

\subsection{Quality Validation of Fabricated Video Clips}
We validate each synthesized clip along two target dimensions using CLIP embeddings~\citep{clip}. 
For semantic relevance, we compute cross-modal cosine similarity between each clip and its corresponding event description, regenerating clips below 0.30 until the threshold is met. 
For scene inconsistency, we compute intra-video visual similarity between each synthesized clip and the remaining real clips, regenerating clips above 0.70. 
This iterative generation enforces both intended properties, yielding clips semantically meaningful yet visually distinguishable.

\subsection{Detailed Descriptions of LMMs}
\label{app:Detailed Descriptions of LMMs}
\textbf{LLaVA-OneVision}~\citep{llavaov} is a model consolidating insights from the LLaVA-NeXT series, built on Qwen-2~\citep{qwen2} and SigLIP~\citep{siglip} vision encoder. 
It employs a Higher AnyRes strategy with balanced token allocation to flexibly represent visual signals across single-image, multi-image, and video scenarios. 
The model follows a curriculum learning process consisting of language-image alignment, high-quality knowledge learning, and a OneVision stage blending 1.6M mixed-modality samples, enabling strong task transfer from images to videos.

\textbf{Qwen3-VL}~\citep{qwen3vl} is built on Qwen3~\citep{qwen3} and SigLIP-2 encoder~\citep{siglip2}, supporting 256K-token interleaved contexts. 
With Interleaved MROPE, DeepStack, and text timestamps, it achieves leading results on reasoning benchmarks.

\textbf{InternVL3.5}~\citep{internvl35} is built on Qwen3 and InternViT vision encoder, proposing Cascade RL that combines offline and online reinforcement learning to enhance reasoning. 
With ViR dynamically adjusting visual token resolution and DvD decoupling vision-language deployment, it achieves 4.05× inference speedup.

\textbf{MiMo-VL-SFT}~\citep{mimovl} is built on Qwen2.5-ViT~\citep{qwen25vl} and MiMo-7B-Base~\citep{mimo} (Xiaomi, 2025), undergoing four-stage pre-training with 2.4T tokens. 
Incorporating long Chain-of-Thought reasoning data into late pre-training stages.

\textbf{GPT-5.1}~\citep{gpt5} is OpenAI‘s unified multimodal system featuring a fast main model and a deep reasoning model with real-time routing. Supporting text, image inputs, and tool use, it employs safe-completions safety training. 
It significantly outperforms predecessors on HealthBench with 26-65\% hallucination reduction.

\textbf{Video-R1}~\citep{videor1} is built on Qwen2.5-VL-7B, pioneering the extension of R1 reinforcement learning to video reasoning. 
It proposes T-GRPO to encourage temporal modeling by comparing ordered and shuffled frame sequences.

\textbf{VideoChat-R1.5}~\citep{videochatr15} is built on Qwen2.5-VL, proposing Visual Test-Time Scaling that refines spatio-temporal focus through iterative perception. 
Using GRPO-based reinforcement learning to optimize perception policies, it achieves over 5\% average improvement across 15+ benchmarks on video QA and spatio-temporal perception.

\textbf{Keye-VL-1.5}~\citep{videochatr15} is built on Qwen3-8B and SigLIP, proposing Slow-Fast video encoding that dynamically allocates resolution based on inter-frame similarity. Through four-stage progressive pre-training extending context from 8K to 128K, it outperforms most comparable models.

\textbf{LongVILA-R1}~\citep{longvilar1} is built on LongVILA~\citep{longvila}, pioneering the extension of reinforcement learning to long video reasoning. 
With a 104K long video QA dataset and two-stage training (CoT-SFT + GRPO), and MR-SP system supporting 8,192 frames, it achieves 65.1\% on VideoMME~\citep{videomme}.

\textbf{GLM-4.1V-Thinking}~\citep{glm41vthinking} is built on GLM-4-9B~\citep{glm4} and AIMv2-Huge vision encoder~\citep{aim}, proposing RLCS to enhance multimodal reasoning. 
With cross-domain joint training across STEM, video, and GUI agents, it outperforms Qwen2.5-VL-72B on 29 benchmarks while maintaining 9B scale.

\textbf{Qwen3-VL-Thinking-8B}~\citep{qwen3vl} is built on Qwen3 and SigLIP-2 encoder, enhancing reasoning through RL following long CoT cold-start. 
With Interleaved MROPE and DeepStack architecture, it significantly outperforms the base instruct version on reasoning benchmarks like MathVista~\citep{mathvista}.

\textbf{OneThinker}~\citep{onethinker} is built on Qwen3-VL-Instruct-8B, unifying 10 visual tasks including QA, captioning, spatio-temporal grounding, tracking, and segmentation. 
With OneThinker-600k corpus and EMA-GRPO addressing reward heterogeneity in multi-task RL, it achieves strong performance across 31 benchmarks while enabling cross-task knowledge transfer.

\textbf{MiMo-VL-RL}~\citep{mimovl} is built on Qwen2.5-ViT and MiMo-7B-Base, further optimized from SFT via Mixed On-policy Reinforcement Learning (MORL). 
With joint RLVR and RLHF training integrating diverse reward signals, it achieves 66.7\% on MMMU and 59.4\% on OlympiadBench.

\textbf{GPT-5.1-Thinking}~\citep{gpt5} is trained via reinforcement learning to produce long internal chains of thought before answering, refining its reasoning and recognizing mistakes. Supporting parallel test-time compute with browsing, coding, and tool use, it achieves 46.2\% on HealthBench Hard~\citep{healthbench}.

\begin{figure*}[t]
    \centering
    \includegraphics[width= \linewidth]{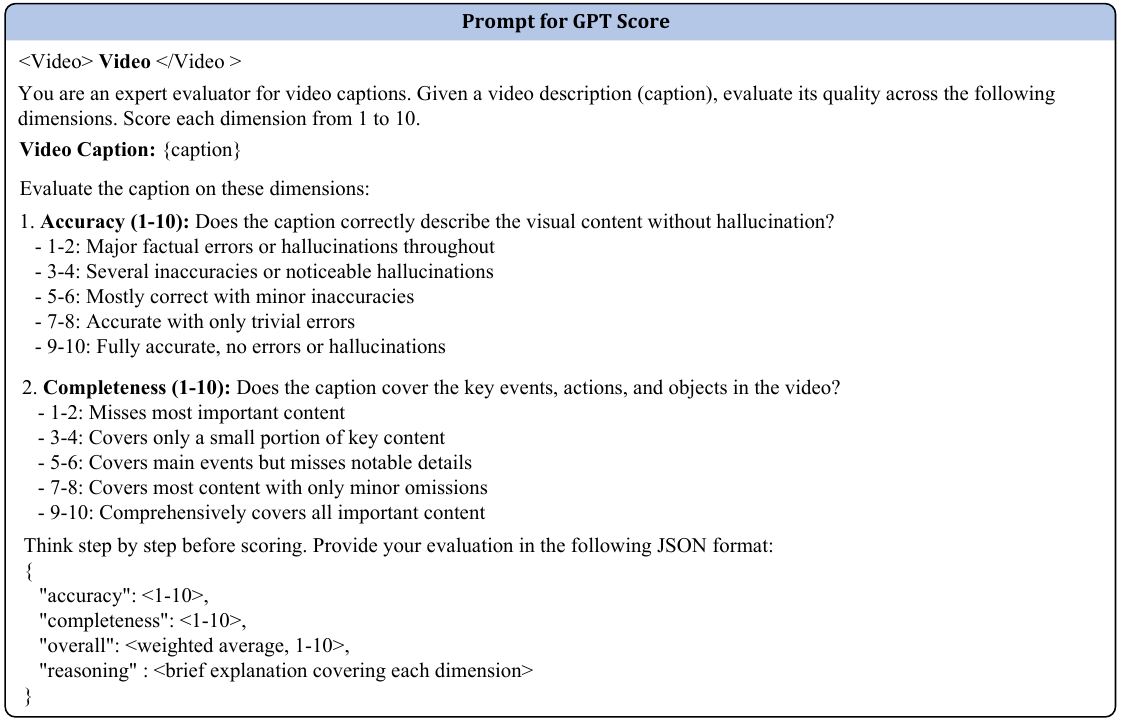}
    \caption{Prompt for GPT Score}
    \label{fig:prompt}
\end{figure*}

\section{Anti-Distillation Policy Optimization}
\label{app:adpo}
\subsection{Implementation Details}
All experiments are conducted on 8 NVIDIA A100 GPUs with a global batch size of 128 with learning rate of $1\mathrm{e}{-6}$.
Models are trained for 1 epoch on 20K QA pairs sampled from Video-R1-260k~\citep{videor1}.
Videos are uniformly sampled to 16 frames during training and 128 frames during evaluation. 
The group size for rollout is set to $G=5$, and the anti-distillation coefficient is set to $\lambda=0.01$. 
All video corruption ratios are set to 0.2.
All other hyperparameters follow the default GRPO.
The reported results are averaged over 3 independent runs with different random seeds.

\subsection{Gradient Analysis}
\label{sec:gradient_analysis}

A natural concern with the joint objective in Eq.~\ref{eq:adpo} is whether the GRPO advantage-weighted gradient and the perception-sensitivity regularizer may push the policy in conflicting directions. We provide a gradient-level analysis to characterize their interaction and identify when conflicts may arise.

\textbf{Gradient Decomposition.} Taking the gradient of $J_{\text{ADPO}}(\theta)$ with respect to $\theta$ yields:
\begin{equation}
\nabla_\theta J_{\text{ADPO}} = \underbrace{\mathbb{E}\left[\sum_{i,t} A_i \, \nabla_\theta \log \pi_\theta(o_{i,t} \mid V, q, o_{i,<t})\right]}_{\nabla_\theta J_{\text{GRPO}}} + \lambda \underbrace{\nabla_\theta \mathbb{E}\left[D_{\text{KL}}\!\left(\pi_\theta^{(i,t)} \,\Vert\, \mathrm{sg}(\pi_\theta^{\text{pert},(i,t)})\right)\right]}_{\nabla_\theta \mathcal{L}_{\text{prcp}}}.
\label{eq:adpo_gradient}
\end{equation}
The stop-gradient operator $\mathrm{sg}(\cdot)$ ensures that the regularizer's gradient flows only through the visually-grounded distribution $\pi_\theta(\cdot \mid V, q, o_{i,<t})$, explicitly pushing it away from the prior-dominated counterpart $\pi_\theta(\cdot \mid \tilde{V}, q, o_{i,<t})$.

\textbf{Case-wise Interaction Analysis.} 
Whether the two gradient terms cooperate or conflict depends on the source of the rollout's correctness. 
We enumerate four canonical cases in Table~\ref{tab:gradient_cases}, characterized by (i) whether the rollout is supported by visual evidence or driven by textual priors, and (ii) the sign of the advantage $A_i$.

\begin{table}[h]
\centering
\small
\caption{Interaction between GRPO and perception-sensitivity gradients across rollout types.}
\label{tab:gradient_cases}
\begin{tabular}{clccc}
\toprule
Case & Rollout source & $A_i$ & GRPO direction & $\mathcal{L}_{\text{prcp}}$ direction \\
\midrule
(a) & Visually-grounded, correct & $+$ & Increase $\pi_\theta(o \mid V)$ & Push away from prior \\
(b) & Visually-grounded, incorrect & $-$ & Decrease $\pi_\theta(o \mid V)$ & Push away from prior \\
(c) & Prior-driven, correct (coincidental) & $+$ & Increase $\pi_\theta(o \mid V)$ & Push away from prior \\
(d) & Prior-driven, incorrect & $-$ & Decrease $\pi_\theta(o \mid V)$ & Push away from prior \\
\bottomrule
\end{tabular}
\end{table}

In cases (a), (b), and (d), the two gradients are mutually compatible, the regularizer reinforces the GRPO signal by either amplifying visually-grounded behaviors or suppressing prior-driven ones. 
The only potentially conflicting case is (c), where a rollout happens to be correct due to alignment between textual priors and the ground truth, and the GRPO term reinforces this rollout while $\mathcal{L}_{\text{prcp}}$ pushes the distribution away from the prior. 
We argue that this conflict is benign for the reason of distinct gradient subspaces: The GRPO gradient operates on the log-probability of a specific token $o_{i,t}$, increasing its probability mass locally. 
In contrast, $\mathcal{L}_{\text{prcp}}$ operates on the shape of the full distribution $\pi_\theta(\cdot \mid V, q, o_{i,<t})$, redistributing mass across the vocabulary. 
The two gradients therefore live in largely orthogonal subspaces, $\mathcal{L}_{\text{prcp}}$ does not directly suppress the probability of the specific correct token in case (c), but rather reshapes the distribution to differ from the prior elsewhere, which is compatible with concentrating mass on visually-supported tokens.

\subsection{Sensitivity to the Anti-Distillation Coefficient \texorpdfstring{$\lambda$}{λ}}
\label{sec:lambda_sensitivity}

A central claim of ADPO is that it breaks the accuracy-robustness trade-off observed in GRPO and DAPO, where higher \texttt{Veridicality} R-Avg is consistently accompanied by elevated FR and NFR (shown in Table~\ref{tab:evaluation}). 
To verify this claim, we conduct a sensitivity analysis over the anti-distillation coefficient $\lambda$. 
We train Qwen3-VL-8B with frame shuffling perturbation under $\lambda \in \{0, 0.001, 0.005, 0.01, 0.02, 0.03, 0.04\}$, where $\lambda = 0$ recovers vanilla GRPO. 
All other hyperparameters are kept identical to the main experiments.

\begin{table}[h]
\centering
\small
\caption{Sensitivity of ADPO to the anti-distillation coefficient $\lambda$ on Qwen3-VL-8B. $\lambda = 0$ corresponds to vanilla GRPO. The shaded region marks the regime where both \texttt{Veridicality} accuracy and perturbation robustness are simultaneously improved over GRPO.}
\label{tab:lambda_sensitivity}
\begin{tabular}{cccccc}
\toprule
\multirow{2.5}{*}{$\lambda$} & \multicolumn{1}{c}{\texttt{Veridicality}} & \multicolumn{1}{c}{\texttt{Plausibility}} & \multicolumn{1}{c}{\texttt{Incongruence}} & \multicolumn{1}{c}{\texttt{Deprivation}} & \multirow{2.5}{*}{Regime} \\
\cmidrule(lr){2-2} \cmidrule(lr){3-3} \cmidrule(lr){4-4} \cmidrule(lr){5-5}
& R-Avg $\uparrow$ & FR $\downarrow$ & NFR $\downarrow$ & NFR $\downarrow$ &  \\
\midrule
$0$ (GRPO) & $41.3$ & $52.1$  & $60.6$ & $55.9$ & Baseline \\
$0.001$ & $41.7$ & $50.4$  & $58.3$ & $54.1$ & Under-reg. \\
\rowcolor{gray!15}
$0.005$ & $42.6$ & $44.8$  & $50.1$ & $46.3$ & Sweet spot \\
\rowcolor{gray!15}
$0.01$ & $\mathbf{43.1}$ & $\mathbf{42.2}$  & $\mathbf{47.0}$ & $\mathbf{43.1}$ & Sweet spot \\
\rowcolor{gray!15}
$0.02$ & $42.4$ & $43.5$  & $48.6$ & $44.8$ & Sweet spot \\
$0.03$ & $40.8$ & $46.7$  & $51.9$ & $48.7$ & Over-reg. \\
$0.04$ & $36.2$ & $50.3$  & $56.4$ & $53.2$ & Collapse \\
\bottomrule
\end{tabular}
\end{table}

\textbf{Three Regimes of $\lambda$.} The results in Table~\ref{tab:lambda_sensitivity} reveal three distinct regimes:

\begin{itemize}[leftmargin=*]
    \item \textbf{(i) Under-regularization ($\lambda \leq 0.001$).} The regularizer is too weak to meaningfully reshape the policy, and ADPO's behavior closely tracks GRPO. The trade-off between \texttt{Veridicality} accuracy and perturbation robustness remains visible.
    \item \textbf{(ii) Sweet spot ($\lambda \in [0.005, 0.02]$).} Within this range, ADPO simultaneously improves \texttt{Veridicality} R-Avg over GRPO and reduces FR / NFR. 
    The trade-off between accuracy and robustness is genuinely broken, validating our central claim.
    \item \textbf{(iii) Over-regularization ($\lambda \geq 0.03$).} The regularizer begins to dominate the GRPO objective, indiscriminately pushing the policy away from any prior-aligned behavior, including the case (c) rollouts discussed in Section~\ref{sec:gradient_analysis}. 
    This causes \texttt{Veridicality} accuracy to degrade below the GRPO baseline, while robustness gains also erode because the policy becomes unstable. At $\lambda = 0.04$, performance collapses toward or below baseline across all configurations.
     
\end{itemize}

%% file: natbib.bib
@article{gpt5,
  author       = {OpenAI},
  title        = {OpenAI {GPT-5} System Card},
  journal      = {CoRR},
  volume       = {abs/2601.03267},
  year         = {2026},
  url          = {https://doi.org/10.48550/arXiv.2601.03267},
  doi          = {10.48550/ARXIV.2601.03267},
  eprinttype   = {arXiv},
  eprint       = {2601.03267},
  bibsource    = {dblp computer science bibliography, https://dblp.org}
}

@article{veo3,
  title={Video models are zero-shot learners and reasoners},
  author={Wiedemer, Thadd{\"a}us and Li, Yuxuan and Vicol, Paul and Gu, Shixiang Shane and Matarese, Nick and Swersky, Kevin and Kim, Been and Jaini, Priyank and Geirhos, Robert},
  journal={arXiv preprint arXiv:2509.20328},
  year={2025}
}

@article{mecd+,
  title={MECD+: Unlocking Event-Level Causal Graph Discovery for Video Reasoning},
  author={Chen, Tieyuan and Liu, Huabin and Wang, Yi and Chen, Yihang and He, Tianyao and Gan, Chaofan and He, Huanyu and Lin, Weiyao},
  journal={IEEE Transactions on Pattern Analysis and Machine Intelligence},
  year={2025},
  publisher={IEEE}
}

@article{llavaov,
  author       = {Bo Li and
                  Yuanhan Zhang and
                  Dong Guo and
                  Renrui Zhang and
                  Feng Li and
                  Hao Zhang and
                  Kaichen Zhang and
                  Peiyuan Zhang and
                  Yanwei Li and
                  Ziwei Liu and
                  Chunyuan Li},
  title        = {LLaVA-OneVision: Easy Visual Task Transfer},
  journal      = {Trans. Mach. Learn. Res.},
  volume       = {2025},
  year         = {2025},
  url          = {https://openreview.net/forum?id=zKv8qULV6n},
  bibsource    = {dblp computer science bibliography, https://dblp.org}
}

@article{qwen3vl,
  title={Qwen3-vl technical report},
  author={Bai, Shuai and Cai, Yuxuan and Chen, Ruizhe and Chen, Keqin and Chen, Xionghui and Cheng, Zesen and Deng, Lianghao and Ding, Wei and Gao, Chang and Ge, Chunjiang and others},
  journal={arXiv preprint arXiv:2511.21631},
  year={2025}
}

@article{internvl35,
  title={Internvl3. 5: Advancing open-source multimodal models in versatility, reasoning, and efficiency},
  author={Wang, Weiyun and Gao, Zhangwei and Gu, Lixin and Pu, Hengjun and Cui, Long and Wei, Xingguang and Liu, Zhaoyang and Jing, Linglin and Ye, Shenglong and Shao, Jie and others},
  journal={arXiv preprint arXiv:2508.18265},
  year={2025}
}

@article{mimovl,
  author       = {Zihao Yue and
                  Zhenru Lin and
                  Yifan Song and
                  Weikun Wang and
                  Shuhuai Ren and
                  Shuhao Gu and
                  Shicheng Li and
                  Peidian Li and
                  Liang Zhao and
                  Lei Li and
                  Kainan Bao and
                  Hao Tian and
                  Hailin Zhang and
                  Xiao{-}Gang Wang and
                  Dawei Zhu and
                  Cici and
                  Chenhong He and
                  Bowen Ye and
                  Bowen Shen and
                  Zihan Zhang and
                  Zihan Jiang and
                  Zhixian Zheng and
                  Zhichao Song and
                  Zhenbo Luo and
                  Yue Yu and
                  Yudong Wang and
                  Yuanyuan Tian and
                  Yu Tu and
                  Yihan Yan and
                  Yi Huang and
                  Xu Wang and
                  Xinzhe Xu and
                  Xingchen Song and
                  Xing Zhang and
                  Xing Yong and
                  Xin Zhang and
                  Xiangwei Deng and
                  Wenyu Yang and
                  Wenhan Ma and
                  Weiwei Lv and
                  Weiji Zhuang and
                  Wei Liu and
                  Sirui Deng and
                  Shuo Liu and
                  Shimao Chen and
                  Shihua Yu and
                  Shaohui Liu and
                  Shande Wang and
                  Rui Ma and
                  Qiantong Wang and
                  Peng Wang and
                  Nuo Chen and
                  Menghang Zhu and
                  Kangyang Zhou and
                  Kang Zhou and
                  Kai Fang and
                  Jun Shi and
                  Jinhao Dong and
                  Jiebao Xiao and
                  Jiaming Xu and
                  Huaqiu Liu and
                  Hongshen Xu and
                  Heng Qu and
                  Haochen Zhao and
                  Hanglong Lv and
                  Guoan Wang and
                  Duo Zhang and
                  Dong Zhang and
                  Di Zhang and
                  Chong Ma and
                  Chang Liu and
                  Can Cai and
                  Bingquan Xia},
  title        = {MiMo-VL Technical Report},
  journal      = {CoRR},
  volume       = {abs/2506.03569},
  year         = {2025},
  url          = {https://doi.org/10.48550/arXiv.2506.03569},
  doi          = {10.48550/ARXIV.2506.03569},
  eprinttype   = {arXiv},
  eprint       = {2506.03569},
  bibsource    = {dblp computer science bibliography, https://dblp.org}
}

@article{videor1,
  title={Video-r1: Reinforcing video reasoning in mllms},
  author={Feng, Kaituo and Gong, Kaixiong and Li, Bohao and Guo, Zonghao and Wang, Yibing and Peng, Tianshuo and Wu, Junfei and Zhang, Xiaoying and Wang, Benyou and Yue, Xiangyu},
  journal={arXiv preprint arXiv:2503.21776},
  year={2025}
}

@article{videochatr15,
  title={Videochat-r1. 5: Visual test-time scaling to reinforce multimodal reasoning by iterative perception},
  author={Yan, Ziang and Li, Xinhao and He, Yinan and Yue, Zhengrong and Zeng, Xiangyu and Wang, Yali and Qiao, Yu and Wang, Limin and Wang, Yi},
  journal={arXiv preprint arXiv:2509.21100},
  year={2025}
}

@article{keye1.5,
  title={Kwai keye-vl 1.5 technical report},
  author={Yang, Biao and Wen, Bin and Ding, Boyang and Liu, Changyi and Chu, Chenglong and Song, Chengru and Rao, Chongling and Yi, Chuan and Li, Da and Zang, Dunju and others},
  journal={arXiv preprint arXiv:2509.01563},
  year={2025}
}

@article{longvilar1,
  title={Scaling rl to long videos},
  author={Chen, Yukang and Huang, Wei and Shi, Baifeng and Hu, Qinghao and Ye, Hanrong and Zhu, Ligeng and Liu, Zhijian and Molchanov, Pavlo and Kautz, Jan and Qi, Xiaojuan and others},
  journal={arXiv preprint arXiv:2507.07966},
  year={2025}
}

@article{glm41vthinking,
  title={Glm-4.1 v-thinking: Towards versatile multimodal reasoning with scalable reinforcement learning},
  author={Hong, Wenyi and Yu, Wenmeng and Gu, Xiaotao and Wang, Guo and Gan, Guobing and Tang, Haomiao and Cheng, Jiale and Qi, Ji and Ji, Junhui and Pan, Lihang and others},
  journal={arXiv e-prints},
  pages={arXiv--2507},
  year={2025}
}

@article{r1ov,
  author       = {Yi Yang and
                  Xiaoxuan He and
                  Hongkun Pan and
                  Xiyan Jiang and
                  Yan Deng and
                  Xingtao Yang and
                  Haoyu Lu and
                  Dacheng Yin and
                  Fengyun Rao and
                  Minfeng Zhu and
                  Bo Zhang and
                  Wei Chen},
  title        = {R1-Onevision: Advancing Generalized Multimodal Reasoning through Cross-Modal
                  Formalization},
  journal      = {CoRR},
  volume       = {abs/2503.10615},
  year         = {2025},
  url          = {https://doi.org/10.48550/arXiv.2503.10615},
  doi          = {10.48550/ARXIV.2503.10615},
  eprinttype   = {arXiv},
  eprint       = {2503.10615},
  bibsource    = {dblp computer science bibliography, https://dblp.org}
}

@article{glm41v,
  title={Chatglm: A family of large language models from glm-130b to glm-4 all tools},
  author={Glm, Team and Zeng, Aohan and Xu, Bin and Wang, Bowen and Zhang, Chenhui and Yin, Da and Zhang, Dan and Rojas, Diego and Feng, Guanyu and Zhao, Hanlin and others},
  journal={arXiv preprint arXiv:2406.12793},
  year={2024}
}

@article{onethinker,
  title={Onethinker: All-in-one reasoning model for image and video},
  author={Feng, Kaituo and Zhang, Manyuan and Li, Hongyu and Fan, Kaixuan and Chen, Shuang and Jiang, Yilei and Zheng, Dian and Sun, Peiwen and Zhang, Yiyuan and Sun, Haoze and others},
  journal={arXiv preprint arXiv:2512.03043},
  year={2025}
}

@article{grpo,
  author       = {DeepSeek{-}AI},
  title        = {DeepSeek-R1: Incentivizing Reasoning Capability in LLMs via Reinforcement
                  Learning},
  journal      = {CoRR},
  volume       = {abs/2501.12948},
  year         = {2025},
  url          = {https://doi.org/10.48550/arXiv.2501.12948},
  doi          = {10.48550/ARXIV.2501.12948},
  eprinttype   = {arXiv},
  eprint       = {2501.12948},
  bibsource    = {dblp computer science bibliography, https://dblp.org}
}

@article{kinetics,
  author       = {Will Kay and
                  Jo{\~{a}}o Carreira and
                  Karen Simonyan and
                  Brian Zhang and
                  Chloe Hillier and
                  Sudheendra Vijayanarasimhan and
                  Fabio Viola and
                  Tim Green and
                  Trevor Back and
                  Paul Natsev and
                  Mustafa Suleyman and
                  Andrew Zisserman},
  title        = {The Kinetics Human Action Video Dataset},
  journal      = {CoRR},
  volume       = {abs/1705.06950},
  year         = {2017},
  url          = {http://arxiv.org/abs/1705.06950},
  eprinttype   = {arXiv},
  eprint       = {1705.06950},
  bibsource    = {dblp computer science bibliography, https://dblp.org}
}

@inproceedings{nextqa,
  author       = {Junbin Xiao and
                  Xindi Shang and
                  Angela Yao and
                  Tat{-}Seng Chua},
  title        = {NExT-QA: Next Phase of Question-Answering to Explaining Temporal Actions},
  booktitle    = {{IEEE} Conference on Computer Vision and Pattern Recognition, {CVPR}
                  2021, virtual, June 19-25, 2021},
  pages        = {9777--9786},
  publisher    = {Computer Vision Foundation / {IEEE}},
  year         = {2021},
  url          = {https://openaccess.thecvf.com/content/CVPR2021/html/Xiao\_NExT-QA\_Next\_Phase\_of\_Question-Answering\_to\_Explaining\_Temporal\_Actions\_CVPR\_2021\_paper.html},
  doi          = {10.1109/CVPR46437.2021.00965},
  bibsource    = {dblp computer science bibliography, https://dblp.org}
}

@inproceedings{msrvtt,
  author       = {Jun Xu and
                  Tao Mei and
                  Ting Yao and
                  Yong Rui},
  title        = {{MSR-VTT:} {A} Large Video Description Dataset for Bridging Video
                  and Language},
  booktitle    = {2016 {IEEE} Conference on Computer Vision and Pattern Recognition,
                  {CVPR} 2016, Las Vegas, NV, USA, June 27-30, 2016},
  pages        = {5288--5296},
  publisher    = {{IEEE} Computer Society},
  year         = {2016},
  url          = {https://doi.org/10.1109/CVPR.2016.571},
  doi          = {10.1109/CVPR.2016.571},
  bibsource    = {dblp computer science bibliography, https://dblp.org}
}

@article{r1reward,
  title={R1-reward: Training multimodal reward model through stable reinforcement learning},
  author={Zhang, Yi-Fan and Lu, Xingyu and Hu, Xiao and Fu, Chaoyou and Wen, Bin and Zhang, Tianke and Liu, Changyi and Jiang, Kaiyu and Chen, Kaibing and Tang, Kaiyu and others},
  journal={arXiv preprint arXiv:2505.02835},
  year={2025}
}

@article{causalvqa,
  author       = {Aaron Foss and
                  Chloe Evans and
                  Sasha Mitts and
                  Koustuv Sinha and
                  Ammar Rizvi and
                  Justine T. Kao},
  title        = {CausalVQA: {A} Physically Grounded Causal Reasoning Benchmark for
                  Video Models},
  journal      = {CoRR},
  volume       = {abs/2506.09943},
  year         = {2025},
  url          = {https://doi.org/10.48550/arXiv.2506.09943},
  doi          = {10.48550/ARXIV.2506.09943},
  eprinttype   = {arXiv},
  eprint       = {2506.09943},
  bibsource    = {dblp computer science bibliography, https://dblp.org}
}

@article{vcrbench,
  author       = {Pritam Sarkar and
                  Ali Etemad},
  title        = {VCRBench: Exploring Long-form Causal Reasoning Capabilities of Large
                  Video Language Models},
  journal      = {CoRR},
  volume       = {abs/2505.08455},
  year         = {2025},
  url          = {https://doi.org/10.48550/arXiv.2505.08455},
  doi          = {10.48550/ARXIV.2505.08455},
  eprinttype   = {arXiv},
  eprint       = {2505.08455},
  bibsource    = {dblp computer science bibliography, https://dblp.org}
}

@inproceedings{causalstep,
  author       = {Xuchen Li and
                  Xuzhao Li and
                  Shiyu Hu and
                  Kaiqi Huang and
                  Wentao Zhang},
  editor       = {Sven Koenig and
                  Chad Jenkins and
                  Matthew E. Taylor},
  title        = {CausalStep: {A} Benchmark for Explicit Stepwise Causal Reasoning in
                  Videos},
  booktitle    = {Fortieth {AAAI} Conference on Artificial Intelligence, Thirty-Eighth
                  Conference on Innovative Applications of Artificial Intelligence,
                  Sixteenth Symposium on Educational Advances in Artificial Intelligence,
                  {AAAI} 2026, Singapore, January 20-27, 2026},
  pages        = {6530--6538},
  publisher    = {{AAAI} Press},
  year         = {2026},
  url          = {https://doi.org/10.1609/aaai.v40i8.37582},
  doi          = {10.1609/AAAI.V40I8.37582},
  bibsource    = {dblp computer science bibliography, https://dblp.org}
}

@inproceedings{clip,
  author       = {Alec Radford and
                  Jong Wook Kim and
                  Chris Hallacy and
                  Aditya Ramesh and
                  Gabriel Goh and
                  Sandhini Agarwal and
                  Girish Sastry and
                  Amanda Askell and
                  Pamela Mishkin and
                  Jack Clark and
                  Gretchen Krueger and
                  Ilya Sutskever},
  editor       = {Marina Meila and
                  Tong Zhang},
  title        = {Learning Transferable Visual Models From Natural Language Supervision},
  booktitle    = {Proceedings of the 38th International Conference on Machine Learning,
                  {ICML} 2021, 18-24 July 2021, Virtual Event},
  series       = {Proceedings of Machine Learning Research},
  pages        = {8748--8763},
  publisher    = {{PMLR}},
  year         = {2021},
  url          = {http://proceedings.mlr.press/v139/radford21a.html},
  bibsource    = {dblp computer science bibliography, https://dblp.org}
}

@article{distill,
  author       = {Geoffrey E. Hinton and
                  Oriol Vinyals and
                  Jeffrey Dean},
  title        = {Distilling the Knowledge in a Neural Network},
  journal      = {CoRR},
  volume       = {abs/1503.02531},
  year         = {2015},
  url          = {http://arxiv.org/abs/1503.02531},
  eprinttype   = {arXiv},
  eprint       = {1503.02531},
  bibsource    = {dblp computer science bibliography, https://dblp.org}
}

@article{dapo,
  author       = {Qiying Yu and
                  Zheng Zhang and
                  Ruofei Zhu and
                  Yufeng Yuan and
                  Xiaochen Zuo and
                  Yu Yue and
                  Tiantian Fan and
                  Gaohong Liu and
                  Lingjun Liu and
                  Xin Liu and
                  Haibin Lin and
                  Zhiqi Lin and
                  Bole Ma and
                  Guangming Sheng and
                  Yuxuan Tong and
                  Chi Zhang and
                  Mofan Zhang and
                  Wang Zhang and
                  Hang Zhu and
                  Jinhua Zhu and
                  Jiaze Chen and
                  Jiangjie Chen and
                  Chengyi Wang and
                  Hongli Yu and
                  Weinan Dai and
                  Yuxuan Song and
                  Xiangpeng Wei and
                  Hao Zhou and
                  Jingjing Liu and
                  Wei{-}Ying Ma and
                  Ya{-}Qin Zhang and
                  Lin Yan and
                  Mu Qiao and
                  Yonghui Wu and
                  Mingxuan Wang},
  title        = {{DAPO:} An Open-Source {LLM} Reinforcement Learning System at Scale},
  journal      = {CoRR},
  volume       = {abs/2503.14476},
  year         = {2025},
  url          = {https://doi.org/10.48550/arXiv.2503.14476},
  doi          = {10.48550/ARXIV.2503.14476},
  eprinttype   = {arXiv},
  eprint       = {2503.14476},
  bibsource    = {dblp computer science bibliography, https://dblp.org}
}

@inproceedings{videomme,
  author       = {Chaoyou Fu and
                  Yuhan Dai and
                  Yongdong Luo and
                  Lei Li and
                  Shuhuai Ren and
                  Renrui Zhang and
                  Zihan Wang and
                  Chenyu Zhou and
                  Yunhang Shen and
                  Mengdan Zhang and
                  Peixian Chen and
                  Yanwei Li and
                  Shaohui Lin and
                  Sirui Zhao and
                  Ke Li and
                  Tong Xu and
                  Xiawu Zheng and
                  Enhong Chen and
                  Caifeng Shan and
                  Ran He and
                  Xing Sun},
  title        = {Video-MME: The First-Ever Comprehensive Evaluation Benchmark of Multi-modal
                  LLMs in Video Analysis},
  booktitle    = {{IEEE/CVF} Conference on Computer Vision and Pattern Recognition,
                  {CVPR} 2025, Nashville, TN, USA, June 11-15, 2025},
  pages        = {24108--24118},
  publisher    = {Computer Vision Foundation / {IEEE}},
  year         = {2025},
  url          = {https://openaccess.thecvf.com/content/CVPR2025/html/Fu\_Video-MME\_The\_First-Ever\_Comprehensive\_Evaluation\_Benchmark\_of\_Multi-modal\_LLMs\_in\_CVPR\_2025\_paper.html},
  doi          = {10.1109/CVPR52734.2025.02245},
  bibsource    = {dblp computer science bibliography, https://dblp.org}
}

@inproceedings{longvideobench,
  author       = {Haoning Wu and
                  Dongxu Li and
                  Bei Chen and
                  Junnan Li},
  editor       = {Amir Globersons and
                  Lester Mackey and
                  Danielle Belgrave and
                  Angela Fan and
                  Ulrich Paquet and
                  Jakub M. Tomczak and
                  Cheng Zhang},
  title        = {LongVideoBench: {A} Benchmark for Long-context Interleaved Video-Language
                  Understanding},
  booktitle    = {Advances in Neural Information Processing Systems 38: Annual Conference
                  on Neural Information Processing Systems 2024, NeurIPS 2024, Vancouver,
                  BC, Canada, December 10 - 15, 2024},
  year         = {2024},
  url          = {http://papers.nips.cc/paper\_files/paper/2024/hash/329ad516cf7a6ac306f29882e9c77558-Abstract-Datasets\_and\_Benchmarks\_Track.html},
  bibsource    = {dblp computer science bibliography, https://dblp.org}
}

@inproceedings{mvbench,
  author       = {Kunchang Li and
                  Yali Wang and
                  Yinan He and
                  Yizhuo Li and
                  Yi Wang and
                  Yi Liu and
                  Zun Wang and
                  Jilan Xu and
                  Guo Chen and
                  Ping Lou and
                  Limin Wang and
                  Yu Qiao},
  title        = {MVBench: {A} Comprehensive Multi-modal Video Understanding Benchmark},
  booktitle    = {{IEEE/CVF} Conference on Computer Vision and Pattern Recognition,
                  {CVPR} 2024, Seattle, WA, USA, June 16-22, 2024},
  pages        = {22195--22206},
  publisher    = {{IEEE}},
  year         = {2024},
  url          = {https://doi.org/10.1109/CVPR52733.2024.02095},
  doi          = {10.1109/CVPR52733.2024.02095},
  bibsource    = {dblp computer science bibliography, https://dblp.org}
}

@article{r1vl,
  author       = {Jingyi Zhang and
                  Jiaxing Huang and
                  Huanjin Yao and
                  Shunyu Liu and
                  Xikun Zhang and
                  Shijian Lu and
                  Dacheng Tao},
  title        = {{R1-VL:} Learning to Reason with Multimodal Large Language Models
                  via Step-wise Group Relative Policy Optimization},
  journal      = {CoRR},
  volume       = {abs/2503.12937},
  year         = {2025},
  url          = {https://doi.org/10.48550/arXiv.2503.12937},
  doi          = {10.48550/ARXIV.2503.12937},
  eprinttype   = {arXiv},
  eprint       = {2503.12937},
  bibsource    = {dblp computer science bibliography, https://dblp.org}
}

@inproceedings{mmvu,
  author       = {Yilun Zhao and
                  Haowei Zhang and
                  Lujing Xie and
                  Tongyan Hu and
                  Guo Gan and
                  Yitao Long and
                  Zhiyuan Hu and
                  Weiyuan Chen and
                  Chuhan Li and
                  Zhijian Xu and
                  Chengye Wang and
                  Ziyao Shangguan and
                  Zhenwen Liang and
                  Yixin Liu and
                  Chen Zhao and
                  Arman Cohan},
  title        = {{MMVU:} Measuring Expert-Level Multi-Discipline Video Understanding},
  booktitle    = {{IEEE/CVF} Conference on Computer Vision and Pattern Recognition,
                  {CVPR} 2025, Nashville, TN, USA, June 11-15, 2025},
  pages        = {8475--8489},
  publisher    = {Computer Vision Foundation / {IEEE}},
  year         = {2025},
  url          = {https://openaccess.thecvf.com/content/CVPR2025/html/Zhao\_MMVU\_Measuring\_Expert-Level\_Multi-Discipline\_Video\_Understanding\_CVPR\_2025\_paper.html},
  doi          = {10.1109/CVPR52734.2025.00793},
  bibsource    = {dblp computer science bibliography, https://dblp.org}
}

@inproceedings{casualvidqa,
  author       = {Jiangtong Li and
                  Li Niu and
                  Liqing Zhang},
  title        = {From Representation to Reasoning: Towards both Evidence and Commonsense
                  Reasoning for Video Question-Answering},
  booktitle    = {{IEEE/CVF} Conference on Computer Vision and Pattern Recognition,
                  {CVPR} 2022, New Orleans, LA, USA, June 18-24, 2022},
  pages        = {21241--21250},
  publisher    = {{IEEE}},
  year         = {2022},
  url          = {https://doi.org/10.1109/CVPR52688.2022.02059},
  doi          = {10.1109/CVPR52688.2022.02059},
  bibsource    = {dblp computer science bibliography, https://dblp.org}
}

@inproceedings{nextgqa,
  author       = {Junbin Xiao and
                  Angela Yao and
                  Yicong Li and
                  Tat{-}Seng Chua},
  title        = {Can {I} Trust Your Answer? Visually Grounded Video Question Answering},
  booktitle    = {{IEEE/CVF} Conference on Computer Vision and Pattern Recognition,
                  {CVPR} 2024, Seattle, WA, USA, June 16-22, 2024},
  pages        = {13204--13214},
  publisher    = {{IEEE}},
  year         = {2024},
  url          = {https://doi.org/10.1109/CVPR52733.2024.01254},
  doi          = {10.1109/CVPR52733.2024.01254},
  bibsource    = {dblp computer science bibliography, https://dblp.org}
}

@inproceedings{siglip,
  author       = {Xiaohua Zhai and
                  Basil Mustafa and
                  Alexander Kolesnikov and
                  Lucas Beyer},
  title        = {Sigmoid Loss for Language Image Pre-Training},
  booktitle    = {{IEEE/CVF} International Conference on Computer Vision, {ICCV} 2023,
                  Paris, France, October 1-6, 2023},
  pages        = {11941--11952},
  publisher    = {{IEEE}},
  year         = {2023},
  url          = {https://doi.org/10.1109/ICCV51070.2023.01100},
  doi          = {10.1109/ICCV51070.2023.01100},
  bibsource    = {dblp computer science bibliography, https://dblp.org}
}

@article{qwen2,
  author       = {An Yang and
                  Baosong Yang and
                  Binyuan Hui and
                  Bo Zheng and
                  Bowen Yu and
                  Chang Zhou and
                  Chengpeng Li and
                  Chengyuan Li and
                  Dayiheng Liu and
                  Fei Huang and
                  Guanting Dong and
                  Haoran Wei and
                  Huan Lin and
                  Jialong Tang and
                  Jialin Wang and
                  Jian Yang and
                  Jianhong Tu and
                  Jianwei Zhang and
                  Jianxin Ma and
                  Jianxin Yang and
                  Jin Xu and
                  Jingren Zhou and
                  Jinze Bai and
                  Jinzheng He and
                  Junyang Lin and
                  Kai Dang and
                  Keming Lu and
                  Keqin Chen and
                  Kexin Yang and
                  Mei Li and
                  Mingfeng Xue and
                  Na Ni and
                  Pei Zhang and
                  Peng Wang and
                  Ru Peng and
                  Rui Men and
                  Ruize Gao and
                  Runji Lin and
                  Shijie Wang and
                  Shuai Bai and
                  Sinan Tan and
                  Tianhang Zhu and
                  Tianhao Li and
                  Tianyu Liu and
                  Wenbin Ge and
                  Xiaodong Deng and
                  Xiaohuan Zhou and
                  Xingzhang Ren and
                  Xinyu Zhang and
                  Xipin Wei and
                  Xuancheng Ren and
                  Xuejing Liu and
                  Yang Fan and
                  Yang Yao and
                  Yichang Zhang and
                  Yu Wan and
                  Yunfei Chu and
                  Yuqiong Liu and
                  Zeyu Cui and
                  Zhenru Zhang and
                  Zhifang Guo and
                  Zhihao Fan},
  title        = {Qwen2 Technical Report},
  journal      = {CoRR},
  volume       = {abs/2407.10671},
  year         = {2024},
  url          = {https://doi.org/10.48550/arXiv.2407.10671},
  doi          = {10.48550/ARXIV.2407.10671},
  eprinttype   = {arXiv},
  eprint       = {2407.10671},
  bibsource    = {dblp computer science bibliography, https://dblp.org}
}

@article{qwen3,
  author       = {Qwen Team},
  title        = {Qwen3 Technical Report},
  journal      = {CoRR},
  volume       = {abs/2505.09388},
  year         = {2025},
  url          = {https://doi.org/10.48550/arXiv.2505.09388},
  doi          = {10.48550/ARXIV.2505.09388},
  eprinttype   = {arXiv},
  eprint       = {2505.09388},
  bibsource    = {dblp computer science bibliography, https://dblp.org}
}

@article{siglip2,
  author       = {Michael Tschannen and
                  Alexey A. Gritsenko and
                  Xiao Wang and
                  Muhammad Ferjad Naeem and
                  Ibrahim Alabdulmohsin and
                  Nikhil Parthasarathy and
                  Talfan Evans and
                  Lucas Beyer and
                  Ye Xia and
                  Basil Mustafa and
                  Olivier J. H{\'{e}}naff and
                  Jeremiah Harmsen and
                  Andreas Steiner and
                  Xiaohua Zhai},
  title        = {SigLIP 2: Multilingual Vision-Language Encoders with Improved Semantic
                  Understanding, Localization, and Dense Features},
  journal      = {CoRR},
  volume       = {abs/2502.14786},
  year         = {2025},
  url          = {https://doi.org/10.48550/arXiv.2502.14786},
  doi          = {10.48550/ARXIV.2502.14786},
  eprinttype   = {arXiv},
  eprint       = {2502.14786},
  bibsource    = {dblp computer science bibliography, https://dblp.org}
}

@article{qwen25vl,
  author       = {Shuai Bai and
                  Keqin Chen and
                  Xuejing Liu and
                  Jialin Wang and
                  Wenbin Ge and
                  Sibo Song and
                  Kai Dang and
                  Peng Wang and
                  Shijie Wang and
                  Jun Tang and
                  Humen Zhong and
                  Yuanzhi Zhu and
                  Ming{-}Hsuan Yang and
                  Zhaohai Li and
                  Jianqiang Wan and
                  Pengfei Wang and
                  Wei Ding and
                  Zheren Fu and
                  Yiheng Xu and
                  Jiabo Ye and
                  Xi Zhang and
                  Tianbao Xie and
                  Zesen Cheng and
                  Hang Zhang and
                  Zhibo Yang and
                  Haiyang Xu and
                  Junyang Lin},
  title        = {Qwen2.5-VL Technical Report},
  journal      = {CoRR},
  volume       = {abs/2502.13923},
  year         = {2025},
  url          = {https://doi.org/10.48550/arXiv.2502.13923},
  doi          = {10.48550/ARXIV.2502.13923},
  eprinttype   = {arXiv},
  eprint       = {2502.13923},
  bibsource    = {dblp computer science bibliography, https://dblp.org}
}

@article{mimo,
  author       = {Bingquan Xia and
                  Bowen Shen and
                  Cici and
                  Dawei Zhu and
                  Di Zhang and
                  Gang Wang and
                  Hailin Zhang and
                  Huaqiu Liu and
                  Jiebao Xiao and
                  Jinhao Dong and
                  Liang Zhao and
                  Peidian Li and
                  Peng Wang and
                  Shihua Yu and
                  Shimao Chen and
                  Weikun Wang and
                  Wenhan Ma and
                  Xiangwei Deng and
                  Yi Huang and
                  Yifan Song and
                  Zihan Jiang and
                  Bowen Ye and
                  Can Cai and
                  Chenhong He and
                  Dong Zhang and
                  Duo Zhang and
                  Guoan Wang and
                  Hao Tian and
                  Haochen Zhao and
                  Heng Qu and
                  Hongshen Xu and
                  Jun Shi and
                  Kainan Bao and
                  QingKai Fang and
                  Kang Zhou and
                  Kangyang Zhou and
                  Lei Li and
                  Menghang Zhu and
                  Nuo Chen and
                  Qiantong Wang and
                  Shaohui Liu and
                  Shicheng Li and
                  Shuhao Gu and
                  Shuhuai Ren and
                  Shuo Liu and
                  Sirui Deng and
                  Weiji Zhuang and
                  Weiwei Lv and
                  Wenyu Yang and
                  Xin Zhang and
                  Xing Yong and
                  Xing Zhang and
                  Xingchen Song and
                  Xinzhe Xu and
                  Xu Wang and
                  Yihan Yan and
                  Yu Tu and
                  Yuanyuan Tian and
                  Yudong Wang and
                  Yue Yu and
                  Zhenru Lin and
                  Zhichao Song and
                  Zihao Yue},
  title        = {MiMo: Unlocking the Reasoning Potential of Language Model - From Pretraining
                  to Posttraining},
  journal      = {CoRR},
  volume       = {abs/2505.07608},
  year         = {2025},
  url          = {https://doi.org/10.48550/arXiv.2505.07608},
  doi          = {10.48550/ARXIV.2505.07608},
  eprinttype   = {arXiv},
  eprint       = {2505.07608},
  bibsource    = {dblp computer science bibliography, https://dblp.org}
}

@article{vlmr1,
  author       = {Wenxuan Huang and
                  Bohan Jia and
                  Zijie Zhai and
                  Shaosheng Cao and
                  Zheyu Ye and
                  Fei Zhao and
                  Zhe Xu and
                  Yao Hu and
                  Shaohui Lin},
  title        = {Vision-R1: Incentivizing Reasoning Capability in Multimodal Large
                  Language Models},
  journal      = {CoRR},
  volume       = {abs/2503.06749},
  year         = {2025},
  url          = {https://doi.org/10.48550/arXiv.2503.06749},
  doi          = {10.48550/ARXIV.2503.06749},
  eprinttype   = {arXiv},
  eprint       = {2503.06749},
  bibsource    = {dblp computer science bibliography, https://dblp.org}
}

@inproceedings{longvila,
  author       = {Yukang Chen and
                  Fuzhao Xue and
                  Dacheng Li and
                  Qinghao Hu and
                  Ligeng Zhu and
                  Xiuyu Li and
                  Yunhao Fang and
                  Haotian Tang and
                  Shang Yang and
                  Zhijian Liu and
                  Yihui He and
                  Hongxu Yin and
                  Pavlo Molchanov and
                  Jan Kautz and
                  Linxi Fan and
                  Yuke Zhu and
                  Yao Lu and
                  Song Han},
  title        = {LongVILA: Scaling Long-Context Visual Language Models for Long Videos},
  booktitle    = {The Thirteenth International Conference on Learning Representations,
                  {ICLR} 2025, Singapore, April 24-28, 2025},
  publisher    = {OpenReview.net},
  year         = {2025},
  url          = {https://openreview.net/forum?id=wCXAlfvCy6},
  bibsource    = {dblp computer science bibliography, https://dblp.org}
}

@article{glm4,
  author       = {Aohan Zeng and
                  Bin Xu and
                  Bowen Wang and
                  Chenhui Zhang and
                  Da Yin and
                  Diego Rojas and
                  Guanyu Feng and
                  Hanlin Zhao and
                  Hanyu Lai and
                  Hao Yu and
                  Hongning Wang and
                  Jiadai Sun and
                  Jiajie Zhang and
                  Jiale Cheng and
                  Jiayi Gui and
                  Jie Tang and
                  Jing Zhang and
                  Juanzi Li and
                  Lei Zhao and
                  Lindong Wu and
                  Lucen Zhong and
                  Mingdao Liu and
                  Minlie Huang and
                  Peng Zhang and
                  Qinkai Zheng and
                  Rui Lu and
                  Shuaiqi Duan and
                  Shudan Zhang and
                  Shulin Cao and
                  Shuxun Yang and
                  Weng Lam Tam and
                  Wenyi Zhao and
                  Xiao Liu and
                  Xiao Xia and
                  Xiaohan Zhang and
                  Xiaotao Gu and
                  Xin Lv and
                  Xinghan Liu and
                  Xinyi Liu and
                  Xinyue Yang and
                  Xixuan Song and
                  Xunkai Zhang and
                  Yifan An and
                  Yifan Xu and
                  Yilin Niu and
                  Yuantao Yang and
                  Yueyan Li and
                  Yushi Bai and
                  Yuxiao Dong and
                  Zehan Qi and
                  Zhaoyu Wang and
                  Zhen Yang and
                  Zhengxiao Du and
                  Zhenyu Hou and
                  Zihan Wang},
  title        = {ChatGLM: {A} Family of Large Language Models from {GLM-130B} to {GLM-4}
                  All Tools},
  journal      = {CoRR},
  volume       = {abs/2406.12793},
  year         = {2024},
  url          = {https://doi.org/10.48550/arXiv.2406.12793},
  doi          = {10.48550/ARXIV.2406.12793},
  eprinttype   = {arXiv},
  eprint       = {2406.12793},
  bibsource    = {dblp computer science bibliography, https://dblp.org}
}

@inproceedings{aim,
  author       = {Enrico Fini and
                  Mustafa Shukor and
                  Xiujun Li and
                  Philipp Dufter and
                  Michal Klein and
                  David Haldimann and
                  Sai Aitharaju and
                  Victor G. Turrisi da Costa and
                  Louis B{\'{e}}thune and
                  Zhe Gan and
                  Alexander Toshev and
                  Marcin Eichner and
                  Moin Nabi and
                  Yinfei Yang and
                  Joshua M. Susskind and
                  Alaaeldin El{-}Nouby},
  title        = {Multimodal Autoregressive Pre-training of Large Vision Encoders},
  booktitle    = {{IEEE/CVF} Conference on Computer Vision and Pattern Recognition,
                  {CVPR} 2025, Nashville, TN, USA, June 11-15, 2025},
  pages        = {9641--9654},
  publisher    = {Computer Vision Foundation / {IEEE}},
  year         = {2025},
  url          = {https://openaccess.thecvf.com/content/CVPR2025/html/Fini\_Multimodal\_Autoregressive\_Pre-training\_of\_Large\_Vision\_Encoders\_CVPR\_2025\_paper.html},
  doi          = {10.1109/CVPR52734.2025.00901},
  bibsource    = {dblp computer science bibliography, https://dblp.org}
}

@article{mathvista,
  title={Mathvista: Evaluating mathematical reasoning of foundation models in visual contexts},
  author={Lu, Pan and Bansal, Hritik and Xia, Tony and Liu, Jiacheng and Li, Chunyuan and Hajishirzi, Hannaneh and Cheng, Hao and Chang, Kai-Wei and Galley, Michel and Gao, Jianfeng},
  journal={arXiv preprint arXiv:2310.02255},
  year={2023}
}

@article{healthbench,
  title={Healthbench: Evaluating large language models towards improved human health},
  author={Arora, Rahul K and Wei, Jason and Hicks, Rebecca Soskin and Bowman, Preston and Qui{\~n}onero-Candela, Joaquin and Tsimpourlas, Foivos and Sharman, Michael and Shah, Meghan and Vallone, Andrea and Beutel, Alex and others},
  journal={arXiv preprint arXiv:2505.08775},
  year={2025}
}

@article{visionr1,
  author       = {Wenxuan Huang and
                  Bohan Jia and
                  Zijie Zhai and
                  Shaosheng Cao and
                  Zheyu Ye and
                  Fei Zhao and
                  Zhe Xu and
                  Yao Hu and
                  Shaohui Lin},
  title        = {Vision-R1: Incentivizing Reasoning Capability in Multimodal Large
                  Language Models},
  journal      = {CoRR},
  volume       = {abs/2503.06749},
  year         = {2025},
  url          = {https://doi.org/10.48550/arXiv.2503.06749},
  doi          = {10.48550/ARXIV.2503.06749},
  eprinttype   = {arXiv},
  eprint       = {2503.06749},
  bibsource    = {dblp computer science bibliography, https://dblp.org}
}
